%% file: main.tex
\pdfoutput=1
\documentclass[10pt,twoside]{article}
\usepackage[letterpaper,textwidth=6.75in,textheight=9.25in,columnsep=0.25in,centering]{geometry}
\usepackage{amsmath}
\usepackage{microtype}
\usepackage[round]{natbib}

\usepackage[hyphens]{url}
\usepackage{graphicx}
\usepackage{booktabs}
\usepackage{amssymb}
\usepackage{amsfonts}
\usepackage{amsthm}
\usepackage{xcolor}
\definecolor{rankred}{HTML}{FF0000}
\definecolor{rankpurple}{HTML}{800080}
\definecolor{figgray}{HTML}{6B7280}
\definecolor{figorange}{HTML}{B45309}
\definecolor{figblue}{HTML}{1F4E79}
\newcommand{\rankfirst}[1]{\textcolor{rankred}{\textbf{#1}}}
\newcommand{\ranksecond}[1]{\textcolor{rankpurple}{\textbf{#1}}}
\newcommand{\rankthird}[1]{\textbf{#1}}
\newcommand{\crossmean}{\textsc{CrossMean}}
\usepackage{multirow}
\usepackage{placeins}

\theoremstyle{definition}

\newcommand{\bX}{\mathbf{X}}
\newcommand{\bA}{\mathbf{A}}
\newcommand{\bB}{\mathbf{B}}
\newcommand{\bC}{\mathbf{C}}
\newcommand{\bD}{\mathbf{D}}
\newcommand{\bG}{\mathbf{G}}
\newcommand{\bI}{\mathbf{I}}
\newcommand{\bW}{\mathbf{W}}
\newcommand{\bF}{\mathbf{F}}
\newcommand{\R}{\mathbb{R}}
\newcommand{\alignop}{\rho}
\usepackage[hidelinks]{hyperref}
\usepackage[capitalise,noabbrev,nameinlink]{cleveref}
\hypersetup{pdftitle={CrossMean: Model Merging by Pairing Merged and Expert Activations},pdfauthor={Moshe Eliasof}}

\begin{document}

\twocolumn[
\begin{center}
{\LARGE\bfseries CrossMean: Model Merging by Pairing Merged and Expert Activations\par}
\vspace{1.2em}
{\large Moshe Eliasof\par}
\vspace{0.4em}
{\normalsize Faculty of Computer and Information Science, Ben-Gurion University of the Negev\par}
{\normalsize\texttt{eliasof@bgu.ac.il}\par}
\vspace{1.5em}
\end{center}
]

\begin{abstract}
Model merging combines fine-tuned experts into one network without retraining, and closed-form methods
compute each merged layer from statistics of its input activations. However, these
statistics are measured on a single expert, while in the merged model, each layer receives
activations from the merged layers that precede it. Recent work re-measures these statistics in the
merged model but still forms them from a single network. We formulate closed-form methods as a linear system and show that they use the same
matrix on both sides. We then introduce CrossMean, which replaces this matrix on one side with a bilinear statistic that pairs merged and expert activations. From
the same statistics, we also derive a label-free criterion that indicates the task sets in which accounting for the merged activations improves the merge more. Empirically, task sets with a lower criterion value show a larger improvement. On the eight-task benchmark of each vision backbone that we tested and on BERT encoders, CrossMean
improves on merging with statistics measured on the experts, and with all linear layers merged, it also
improves on re-measuring these statistics in the merged model. These results suggest that merging benefits
from pairing merged and expert activations.
\end{abstract}

\input{sections/introduction}
\input{sections/related_work}
\input{sections/method}
\input{sections/criterion}
\input{sections/experiments}
\input{sections/conclusion}

\bibliography{references}

\clearpage
\appendix
\crefalias{section}{appendix}
\crefalias{subsection}{subappendix}
\crefalias{subsubsection}{subsubappendix}
\onecolumn
\input{sections/appendix}

\end{document}

%% file: sections/introduction.tex
\section{Introduction}
\label{sec:intro}

Fine-tuned checkpoints are abundant, and retraining is expensive. Hence, combining several of these checkpoints, which we call experts, into one network by arithmetic on their weights has become a standard tool
\citep{wortsman2022model,ilharco2023editing,yang2024mergingsurvey}. Among model merging methods, one family is
\emph{closed-form and activation-based}. Specifically, these methods (i)~pass a small set of unlabeled inputs, which we call the \emph{calibration set}, through each
expert; (ii)~summarize the activations; and (iii)~solve for the merged weights in one step. Within this family, RegMean \citep{jin2023regmean} merges each linear layer by weighting the experts with the Gram matrix of the inputs of that layer, and it is our reference method.

\begin{figure*}[t]
\centering
\begin{minipage}[b]{0.25\textwidth}\centering\includegraphics[width=\linewidth]{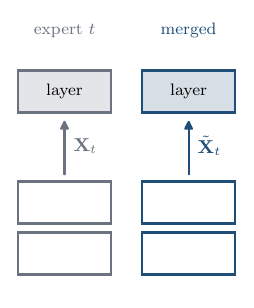}\\[2pt]{\small (a) Expert inputs vs.\ merged inputs}\end{minipage}\hfill
\begin{minipage}[b]{0.27\textwidth}\centering\includegraphics[width=\linewidth]{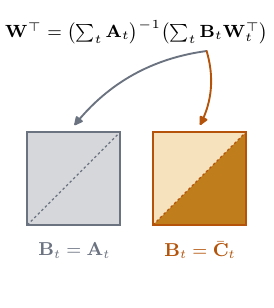}\\[2pt]{\small (b) The matrix $\bB_t$ in \Cref{eq:template}}\end{minipage}\hfill
\begin{minipage}[b]{0.44\textwidth}\centering\includegraphics[width=\linewidth]{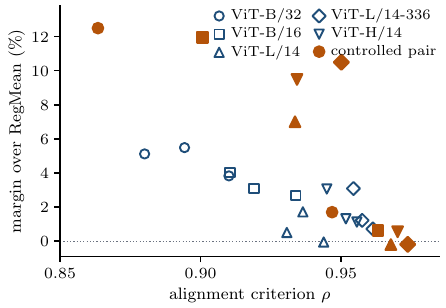}\\[2pt]{\small (c) Margin over RegMean vs.\ $\alignop$}\end{minipage}
\caption{\textbf{(a)}~Existing closed-form methods measure the inputs $\bX_t$ of each layer on expert
$t$, while in the merged model, the layer receives the merged inputs $\tilde\bX_t$. \textbf{(b)}~In the
linear system of \Cref{eq:template}, the existing activation-based methods set $\bB_t=\bA_t$, a symmetric
\textcolor{figgray}{Gram matrix (gray)}, while our \crossmean{} sets $\bB_t=\bar\bC_t$, the
\textcolor{figorange}{bilinear statistic (orange)} of \Cref{eq:bilinear}, which pairs the merged and the
expert inputs. \textbf{(c)}~The margin of our \crossmean{} over RegMean tends to decrease as our alignment criterion $\alignop$ of
\Cref{eq:rho} approaches $1$, both across backbones and numbers of tasks, shown by the open markers, and
with only the tasks varied, shown by the filled markers.}
\label{fig:thesis}
\end{figure*}

However, existing methods such as RegMean and Fisher merging \citep{matena2022merging} fit each layer
using statistics gathered from \emph{one} network. For example, RegMean measures the Gram matrix of the inputs of
each layer while running a single expert, and Fisher merging measures the gradients of each expert at
the weights of that expert. In contrast, in the merged model, the inputs of a layer are produced by the merged layers that precede it, which we call the \emph{merged prefix} of the layer. Hence, these statistics describe the experts rather than the merged model, and we refer to this discrepancy as the \emph{merged-prefix mismatch}. Moreover, as we illustrate in \Cref{fig:thesis}a, each merged layer changes the input of the
next layer. Consequently, this mismatch can propagate to all subsequent layers.

To make the role of these statistics explicit, we formulate existing merging methods in \Cref{sec:onesolve} as a linear system per layer, in which the merged weights are obtained by multiplying the weights of each expert by one matrix and normalizing by the sum, over the experts, of a second matrix. As we summarize in \Cref{tab:onesolve}, all of these methods use the same matrix in both roles, and for the activation-based methods, this matrix is a Gram matrix measured on a single network. 

However, we identify that a merge involves \emph{two} input distributions: (i)~the activations that the layer receives in the merged model, which its merged prefix produces and which we call the \emph{merged inputs}; and (ii)~the activations that the same layer receives in each expert, which we call the \emph{expert inputs}. Specifically, the merged layer should map the merged inputs to the outputs that the expert produces on the expert inputs. In contrast, a Gram matrix describes only one of these two distributions, and hence it does not represent the relation between them.

RegMean++ \citep{nguyen2026regmeanpp} and Chain of Merges \citep{buzzega2025chainofmerges} address the merged-prefix mismatch by merging the layers in depth order and re-measuring their inputs on the merged prefix, at the boundaries between transformer blocks and before every layer, respectively. We refer to this approach as the \emph{depth-ordered correction}, and to its per-layer form \citep{buzzega2025chainofmerges} as \emph{Progressive RegMean}. Progressive RegMean measures the Gram matrix on the merged inputs instead of the expert inputs, but it still uses
the same matrix on both sides of the linear system. Moreover, in our experiments, its improvement over RegMean
varies across task sets, that is, across the sets of tasks whose experts are merged, from a
fraction of a percent to several percent. These observations raise two questions: (i)~how should the linear system of each layer change when it uses both the merged and the expert inputs; and (ii)~can we anticipate from unlabeled data how much accounting for the merged inputs improves on RegMean?

To answer the first question, we keep the two inputs distinct in the linear system itself. As a result, the right-hand side of the normal equations contains the \emph{cross statistic}, which pairs the merged inputs with the expert inputs, in place of the Gram matrix. Averaging the cross statistic along a path of networks from the merged prefix to the expert yields the \emph{bilinear statistic}, on which we build our method, \emph{\crossmean{}} (\textbf{Cross}-statistic Regression \textbf{Mean}), which we describe in \Cref{sec:crossmean} and illustrate in
\Cref{fig:thesis}b.

To answer the second question, we observe that Progressive RegMean and \crossmean{} differ from RegMean only when the merged prefix differs from the experts. We therefore measure the agreement between the two statistics that involve the merged inputs: (i)~the Gram matrix of the merged inputs; and (ii)~the cross statistic. Our \emph{alignment criterion}, presented in \Cref{sec:criterion} and shown in \Cref{fig:thesis}c, has no free parameters, uses no labels, and is computed from the experts and the weight-averaged model, namely their average.

\textbf{Contributions.} The contributions of this work are summarized as follows:
\begin{itemize}
\item We extend the linear-system view of closed-form merging, discussed in \Cref{sec:related}, to a broad set of existing methods, and we show that all of them use the same matrix on both sides, which, for the activation-based methods, is a Gram matrix measured on a single network (\Cref{sec:onesolve}).
\item We introduce \crossmean{}, a closed-form merging method whose bilinear statistic pairs the merged and the expert inputs, and which contains RegMean and Progressive RegMean as special cases
(\Cref{sec:crossmean}).
\item We propose a label-free alignment criterion, computed before the closed-form merge, which is lower where Progressive RegMean and \crossmean{} improve more on RegMean (\Cref{sec:criterion}).
\item We show empirically that \crossmean{} improves on RegMean on the eight-task sets of five vision backbones and on BERT encoders, and that with all linear layers merged, it further improves on the depth-ordered correction on each backbone that we tested, and that task sets with a lower criterion value show a larger improvement (\Cref{sec:experiments}).
\end{itemize}

%% file: sections/related_work.tex
\section{Related Work}
\label{sec:related}

\textbf{Data-free merging in weight space.} Weight averaging \citep{wortsman2022model} and Task
Arithmetic \citep{ilharco2023editing} combine experts by summing their task vectors, the differences between the weights of each expert and the pretrained weights, scaled by a global scalar.
Building on these methods, TIES \citep{yadav2023ties}, DARE \citep{yu2024dare}, Model Stock
\citep{jang2024modelstock}, and Iso-C \citep{marczak2025iso} reduce interference between the task vectors or rescale the sum. Many later methods follow the same approach, and we review them in
\Cref{app:morerelated}. These methods do not use activation statistics, and they form the data-free baselines of our comparison. In our formulation as a linear system, which we present in \Cref{sec:onesolve}, all of these methods
except Iso-C weight the experts by the identity matrix, up to a mask, an entrywise reweighting, or a
scalar.  Hence, these methods treat all input directions of a layer in the same way, and we elaborate on them in \Cref{app:protocol}. Iso-C also relates merging quality to an alignment, which it computes from the weights, whereas our criterion is computed from the activations, as we discuss in \Cref{app:criterion}.

\textbf{Statistics-weighted merging.} Fisher merging \citep{matena2022merging} weights parameters by a
diagonal Fisher estimate, and RegMean \citep{jin2023regmean} computes each merged linear layer in closed form, weighting the experts by the Gram matrix of the inputs of that layer. In both methods, the statistics are measured while running a single
expert. Building on these methods, MaTS \citep{tam2024mats} casts Fisher merging and RegMean as instances
of a common linear system and solves it with conjugate gradient. Our formulation in \Cref{sec:onesolve}
follows this view and extends it to a broad set of existing methods, including the weight-space methods. Unlike Fisher merging and RegMean, WUDI-Merging \citep{cheng2025wudi} and ACE-Merging \citep{acemerging2025} derive the input statistics of each layer from the task vectors and need no data. LOT \citep{sun2025lot} fits each linear
layer to reproduce the features of each expert on the expert inputs. As we summarize in \Cref{tab:onesolve}, LOT uses the same matrices as RegMean. Overall, each of these methods weights the experts by statistics of single experts. In contrast, our \crossmean{} replaces the Gram matrix with a bilinear statistic that pairs the merged and the expert inputs.

\textbf{Test-time and gradient-based merging.} AdaMerging \citep{yang2024adamerging} tunes per-layer
coefficients on unlabeled \emph{test} inputs, E-WEMoE \citep{shen2024wemoe} and Twin-Merging
\citep{lu2024twinmerging} learn routing or per-example weights, and ProDistill \citep{xu2025prodistill}
trains the merge layer by layer through distillation. These methods rely on gradient steps, whereas we study closed-form methods whose statistics are computed on a small unlabeled calibration set.

\textbf{Depth-ordered correction.} RegMean++ \citep{nguyen2026regmeanpp} and Chain of Merges (CoM) \citep{buzzega2025chainofmerges}
independently identify the merged-prefix mismatch of \Cref{sec:intro}, and they merge the layers in depth order, re-measuring the layer inputs on the merged prefix. As we show in \Cref{tab:onesolve}, both methods keep the two matrices of the linear system equal and change \emph{where} the Gram matrix is measured. Distinctively, we replace the Gram matrix on the right-hand side with a bilinear statistic that pairs the merged and the expert inputs. Moreover, our criterion in \Cref{sec:criterion} anticipates how much the depth-ordered correction and \crossmean{} improve on RegMean.

%% file: sections/method.tex
\section{Closed-Form Merging and \crossmean{}}
\label{sec:method}

In this Section, we present our method, \crossmean{}. To this end, after introducing our setup and
notation in \Cref{sec:setup}, we (i)~formulate the existing closed-form merging methods as a linear
system in \Cref{sec:onesolve}, and show that all the existing methods in \Cref{tab:onesolve} use the same matrix on both sides of this linear system; (ii)~formalize, in \Cref{sec:progressive}, the merged-prefix mismatch, and show that the depth-ordered correction of prior work changes where this matrix is measured but keeps it on both sides; and (iii)~replace, in \Cref{sec:crossmean}, the Gram matrix with a bilinear statistic of the merged and the expert inputs, which yields \crossmean{}.

\subsection{Setup and notation}
\label{sec:setup}

Let $\theta_0$ be a pretrained backbone and $\theta_1,\dots,\theta_T$ experts
fine-tuned from it on $T$ tasks, with task vectors $\Delta_t=\theta_t-\theta_0$. For each task $t$, we
assume access to a small set $\mathcal{D}_t$ of \emph{unlabeled} inputs from that task, namely the
calibration set of \Cref{sec:intro}. The closed-form methods pass these inputs through the networks to measure the input
statistics of each layer. We consider two choices of which layers are merged in closed form, which we call \emph{coverages}.
With \emph{MLP coverage}, which we use in the shared-protocol grid of \Cref{sec:experiments}, we merge the two linear layers of
every MLP block in closed form, and we merge the attention projections by averaging the weights of the
experts. With \emph{full coverage}, following RegMean \citep{jin2023regmean} and RegMean++ \citep{nguyen2026regmeanpp}, we also merge the attention input and output projections in closed form, which covers every linear layer of the transformer.

Consider a layer that is merged in closed form, which we call a \emph{mergeable layer}. We denote by
$\bW_t\in\R^{d'\times d}$ the weight of expert $t$ in this layer, and by $\bX_t\in\R^{M\times d}$ the
matrix whose $M$ rows are the token inputs that this layer receives when $\mathcal{D}_t$ is passed through
\emph{expert $t$}. Similarly, we denote by $\tilde\bX_t\in\R^{M\times d}$ the matrix of token inputs that the same layer receives when $\mathcal{D}_t$ is passed through its merged prefix. As in \Cref{sec:intro}, we refer to $\tilde\bX_t$ and $\bX_t$ as the merged inputs and the expert inputs, respectively, and we denote by $\bG_t=\bX_t^\top\bX_t$ the Gram matrix of the expert inputs, which RegMean uses. Because $\tilde\bX_t$ and $\bX_t$ are both measured on $\mathcal{D}_t$, their rows $i$ correspond to the same calibration token.

\subsection{Closed-form merging methods as a linear system}
\label{sec:onesolve}

In this Section, we show that many of the closed-form methods of \Cref{sec:related} can be formulated as a linear system per mergeable layer, of the following form:
\begin{equation}
  \bW^\top = \Big(\textstyle\sum_t \bA_t\Big)^{-1}\Big(\textstyle\sum_t \bB_t\,\bW_t^\top\Big),
  \label{eq:template}
\end{equation}
where $\bW$ is the merged weight, $\bA_t$ is the matrix that weights expert $t$ in the normalization on
the left, and $\bB_t$ is the matrix that multiplies the weights of expert $t$ on the right. Each method
corresponds to (i)~a choice of $\bA_t$ and $\bB_t$; and (ii)~the network on which their statistics are measured, as we summarize in
\Cref{tab:onesolve}. For the weight-space methods, we use task-vector coordinates,
$\Delta^\top=(\sum_t\bA_t)^{-1}\sum_t\bB_t\Delta_t^\top$, where $\Delta$ and $\Delta_t$ are the task vectors of the merged layer and of expert $t$ in this layer, and $\bB_t=\lambda\bA_t$ with a scalar $\lambda$. We describe each method in \Cref{app:protocol}. Overall, our formulation reveals a property that all the existing methods in \Cref{tab:onesolve} share, as we show in its columns $\bA_t$ and $\bB_t$: $\bB_t=\bA_t$, up to a scalar for the weight-space methods.

\begin{table}[t]
\centering\footnotesize
\setlength{\tabcolsep}{2pt}
\caption{Our formulation of merging methods as the linear system in \Cref{eq:template}. All the existing methods in this table put the same matrix on both sides, up to a scalar for the weight-space methods, while our
\crossmean{} does not. Here $\bX_t$ and $\tilde\bX_t$ hold the expert and the merged inputs of a layer, $\bG_t=\bX_t^\top\bX_t$ is the
Gram matrix of the expert inputs, $\mathbf{m}_t$ is the TIES mask of expert $t$, and $\hat\bF_t$ is the Fisher information of expert $t$, of which Fisher merging
keeps the diagonal. The \crossmean{} row shows the bilinear statistic $\bar\bC_t$ of \Cref{eq:bilinear}, which averages the cross statistic of \Cref{eq:crossnormal} along a path.}
\label{tab:onesolve}
\footnotesize
\resizebox{\columnwidth}{!}{%
\begin{tabular}{llll}
\toprule
Method & $\bA_t$ & $\bB_t$ & Measured on \\
\midrule
Weight averaging & $\bI$ & $\bI$ & none \\
Task Arithmetic  & $\bI$ & $\lambda\bI$ & none \\
TIES             & $\mathrm{diag}\,\mathbf{m}_t$ & $\lambda\,\mathrm{diag}\,\mathbf{m}_t$ & none \\
Fisher merging & $\mathrm{diag}\,\hat\bF_t$ & $\mathrm{diag}\,\hat\bF_t$ & expert $t$ \\
RegMean          & $\bG_t$ & $\bG_t$ & expert $t$ \\
LOT & $\bG_t$ & $\bG_t$ & expert $t$ \\
RegMean++/CoM & $\tilde\bX_t^\top\tilde\bX_t$ & $\tilde\bX_t^\top\tilde\bX_t$ & merged prefix \\
\midrule
\textbf{\crossmean{} (Ours)} & $\tilde\bX_t^\top\tilde\bX_t$ & $\bar\bC_t$
                     & \textbf{merged prefix, expert} \\
\bottomrule
\end{tabular}}
\end{table}

\subsection{The merged-prefix mismatch}
\label{sec:progressive}

In this Section, we formalize the merged-prefix mismatch, introduced in \Cref{sec:intro}. We then show that the depth-ordered correction of prior work still uses the same matrix on both sides of \Cref{eq:template},
which motivates our method in \Cref{sec:crossmean}.

Specifically, RegMean chooses $\bW$ as the minimizer of
\[
  \sum_t\lVert\bX_t\bW^\top-\bX_t\bW_t^\top\rVert_F^2 ,
\]
that is, the merged layer should reproduce the output of each expert \emph{on the expert inputs}.
However, in the merged model, the layer receives $\tilde\bX_t$ from its merged prefix rather than $\bX_t$. This difference between $\tilde\bX_t$ and $\bX_t$ is the merged-prefix mismatch at the level of a single layer.
Chain of Merges \citep{buzzega2025chainofmerges} and RegMean++ \citep{nguyen2026regmeanpp} address this mismatch with the depth-ordered correction, which replaces $\bX_t$ with $\tilde\bX_t$ and hence evaluates the objective on the merged inputs. As discussed in \Cref{sec:intro}, we refer to the per-layer form of this correction as Progressive RegMean, and we provide its details in \Cref{app:protocol}. However, Progressive RegMean keeps the same Gram matrix
$\tilde\bX_t^\top\tilde\bX_t$ on both sides of \Cref{eq:template}. Hence, $\bB_t=\bA_t$ still holds, and the
relation between the merged and the expert inputs is not represented.

\subsection{\crossmean{}: Bilinear Statistic Merging}
\label{sec:crossmean}

As discussed in \Cref{sec:progressive}, Progressive RegMean corrects where the Gram matrix is measured, but
it does not represent the relation between the merged and the expert inputs. In this Section, we present our method, \crossmean{}, which represents this relation. To this end, it replaces the Gram matrix on the right-hand side of \Cref{eq:template} with a bilinear statistic of the two inputs, which we define in \Cref{eq:bilinear}.

\textbf{Reproducing the outputs of the experts.} Progressive RegMean minimizes
$\sum_t\lVert\tilde\bX_t\bW^\top-\tilde\bX_t\bW_t^\top\rVert_F^2$. In this objective, the target for expert
$t$ is $\tilde\bX_t\bW_t^\top$, namely the weights of expert $t$ applied to the merged inputs. However, expert $t$ does not receive these inputs. Instead, it produces $\bX_t\bW_t^\top$ on the expert inputs. Hence, the target of Progressive RegMean differs from the output of the expert unless
$\tilde\bX_t=\bX_t$. To lift this restriction, we keep $\tilde\bX_t$ as the input of the merged layer,
and we set the target to the output $\bX_t\bW_t^\top$ that expert $t$ produces on the expert inputs. This choice
yields the following objective for the merged layer:
\begin{equation}
  \min_{\bW}\ \sum_t \big\lVert \tilde\bX_t\bW^\top - \bX_t\bW_t^\top \big\rVert_F^2 ,
  \label{eq:crossobj}
\end{equation}
whose normal equations read:
\begin{equation}
  \Big(\sum_t\tilde\bX_t^\top\tilde\bX_t\Big)\bW^\top=\sum_t\bC_t\,\bW_t^\top ,
  \qquad \bC_t=\tilde\bX_t^\top\bX_t .
  \label{eq:crossnormal}
\end{equation}

\textbf{Properties of the cross statistic.} We note that the matrix $\bC_t$ is a \emph{cross} statistic between the merged and the expert inputs,
that is, a product of two different input matrices rather than a Gram matrix. Therefore, in general, $\bC_t$ is neither symmetric
nor positive semi-definite. However, the linear system does not require these properties. Specifically, only the left-hand side of \Cref{eq:crossnormal} is inverted, and this side is the Gram matrix of the merged inputs, which is positive semi-definite and which we regularize with the same ridge term as RegMean. In contrast, $\bC_t$ only multiplies the expert weights on the
right-hand side. Moreover, the asymmetry of the cross statistic is important for \crossmean{}. In particular, in \Cref{sec:exp-correct}, we show that removing the antisymmetric part of the bilinear statistic in \Cref{eq:bilinear} lowers the accuracy on both backbones that we tested.

\textbf{Averaging along the path from the merged prefix to the expert.} The cross statistic $\bC_t$
in \Cref{eq:crossnormal} pairs the merged inputs, produced by the merged prefix, with the expert inputs, produced by expert $t$. Between these two networks, we define a \emph{path} as follows. Let $\theta_{\mathrm{merged}}$ be the merged prefix of the current layer, and let $\theta_t$ be the corresponding layers of expert $t$. For
$s\in[0,1]$, we define the interpolated network $\theta(s)=\theta_{\mathrm{merged}}+s(\theta_t-\theta_{\mathrm{merged}})$,
and we denote by $\bX_t(s)$ the inputs that the current layer receives when $\mathcal{D}_t$ is passed
through $\theta(s)$. Hence, $\bX_t(0)=\tilde\bX_t$ and $\bX_t(1)=\bX_t$, and the set
$\{\theta(s)\}_{s\in[0,1]}$ is the path from the merged prefix to the expert. Building on the cross
statistic, our \crossmean{} adds one design choice: in the target $\bX_t\bW_t^\top$ of \Cref{eq:crossobj}, it replaces $\bX_t$ with $\bX_t(s)$, and it averages
the resulting objective with equal weights over a finite set $\mathcal{S}$ of points $s$ on this path. This averaging does not change the left-hand side of
\Cref{eq:crossnormal}, and it replaces $\bC_t$ on the right-hand side with the average of $\tilde\bX_t^\top\bX_t(s)$ over $\mathcal{S}$, which we call the
\emph{bilinear statistic}:
\begin{equation}
  \bar\bC_t=\tilde\bX_t^\top\Big(\frac{1}{|\mathcal{S}|}\sum_{s\in\mathcal{S}}\bX_t(s)\Big),
  \label{eq:bilinear}
\end{equation}
so that $\bB_t=\bar\bC_t$ in \Cref{eq:template}. The set $\mathcal{S}=\{1\}$ yields $\bC_t$ itself, and the set $\mathcal{S}=\{0\}$ recovers the Gram matrix of the merged inputs, in which case \crossmean{} reduces to Progressive RegMean. Similarly, if the merged prefix coincides with the corresponding layers of every expert, then $\bX_t(s)=\bX_t$ for every $s$ and every $t$, and $\bar\bC_t=\tilde\bX_t^\top\tilde\bX_t=\bG_t$. Hence, \crossmean{} reduces to RegMean. We use $\mathcal{S}=\{0,\tfrac12,1\}$
in all our experiments. This set contains the two ends of the path, at which $\tilde\bX_t^\top\bX_t(s)$ equals the Gram matrix of the merged inputs and the cross statistic, respectively, together with its midpoint. 

\textbf{Implementation.} Following RegMean, Progressive RegMean and \crossmean{} reduce the off-diagonal entries of their statistics, and all three methods add a ridge term scaled by the mean diagonal of the matrix being inverted, as we describe in \Cref{app:protocol}.

%% file: sections/criterion.tex
\section{The Alignment Criterion}
\label{sec:criterion}

In this Section, we derive a label-free criterion that indicates, before the closed-form merge, how much a merge that accounts for the merged inputs, namely our \crossmean{} of \Cref{sec:crossmean} and the depth-ordered correction of \Cref{sec:progressive}, is likely to improve on RegMean. Both methods differ from RegMean only when the merged prefix differs from the experts. Hence, a label-free measure of this difference identifies the task sets in which their solutions can depart from RegMean, and we examine its association with their improvement in \Cref{sec:exp-criterion}. To this end, we build our criterion from the statistics of the linear system in \Cref{eq:template}.

\textbf{From the linear system to two statistics.} In the linear systems of RegMean \citep{jin2023regmean}, Progressive RegMean \citep{buzzega2025chainofmerges}, and \Cref{eq:crossnormal}, the inputs enter only through products $\bX^\top\bX'$ with $\bX,\bX'\in\{\tilde\bX_t,\bX_t\}$. Consequently, the differences between these linear systems depend on the inputs only through these products. We therefore build our criterion on the two statistics that involve the merged inputs,
\begin{equation}
  \bA = \tilde\bX_t^\top\tilde\bX_t , \qquad \bC = \tilde\bX_t^\top\bX_t ,
  \label{eq:AC}
\end{equation}
namely the Gram matrix of the merged inputs, which Progressive RegMean and \crossmean{} invert, and the cross statistic of \Cref{eq:crossnormal}, from which \crossmean{} builds its bilinear statistic. We omit the index $t$ on $\bA$ and $\bC$ to lighten the notation. When the merged inputs equal the expert inputs,
$\bC=\bA=\bG_t$, and these three linear systems coincide. In contrast, as the merged inputs depart from the expert inputs, $\bC$ departs from $\bA$, and these linear systems differ more.

\textbf{The alignment criterion.} We measure the agreement between $\bA$ and $\bC$ by the cosine of the angle between them in the Frobenius inner product,
\begin{equation}
  \alignop \;=\; \frac{\langle \bA, \bC\rangle}{\lVert\bA\rVert_F\,\lVert\bC\rVert_F}
  \;\in\; [-1,1] ,
  \label{eq:rho}
\end{equation}
and our criterion $\alignop$ is the average of this cosine over the mergeable layers and the experts. This quantity has the form of kernel alignment
\citep{cristianini2001kernel}, applied to the Gram matrix and the cross statistic, and we therefore call
$\alignop$ the \emph{alignment criterion}. By the Cauchy--Schwarz inequality, $\alignop\le 1$, with
equality if and only if $\bC$ is a positive multiple of $\bA$. In particular, $\alignop=1$ when the merged inputs equal the expert inputs. Hence, $\alignop$ measures the merged-prefix mismatch through the angle between $\bC$ and $\bA$, independently of their scale, and lower values of $\alignop$ indicate a larger angle between the two statistics.

\textbf{Computing the criterion before the closed-form merge.} To compute $\alignop$, we measure $\tilde\bX_t$ on the weight-averaged model of \Cref{sec:intro}, which is available before any closed-form merge. As a result, our criterion (i)~depends only on the experts and their calibration sets; (ii)~uses no labels and no test data; and (iii)~adds one forward pass per calibration set to the
forward passes that RegMean already performs. We plot $\alignop$ layer by layer for the five vision backbones of our experiments
in \Cref{fig:rhoprofile} of \Cref{app:criterion}.

\textbf{Comparison with the raw prefix drift.} An alternative measure of the merged-prefix mismatch is the \emph{raw prefix drift}, the relative distance $\lVert\tilde\bX_t-\bX_t\rVert_F/\lVert\bX_t\rVert_F$ between the merged and the expert inputs, for which
\[
  \lVert\tilde\bX_t-\bX_t\rVert_F^2=\mathrm{tr}\,\bA-2\,\mathrm{tr}\,\bC+\mathrm{tr}\,\bG_t .
\]
Hence, the drift depends on the statistics in \Cref{eq:AC} only through their traces, and it weights every input
direction equally, including directions in which the merged inputs carry low energy. In contrast,
$\langle\bA,\bC\rangle$ weights each direction by the energy that the merged inputs place in it, as we
detail in \Cref{app:criterion}. Consistent with this analysis, on the eight-task sets of ViT-B/32,
ViT-B/16, and ViT-L/14, a lower $\alignop$ is associated with a larger improvement of the depth-ordered
correction, whereas the drift does not order these improvements. Overall, we expect the improvements of \crossmean{} and of the depth-ordered correction over RegMean to decrease as $\alignop$ approaches $1$, as we test in \Cref{sec:exp-criterion}.

%% file: sections/experiments.tex
\newpage
\section{Experiments}
\label{sec:experiments}

In this Section, we evaluate our method, \crossmean{}, and our alignment criterion. To this end, we (i)~measure the merging accuracy of \crossmean{} and of the baselines in \Cref{sec:exp-correct}; (ii)~test whether our alignment criterion $\alignop$ in \Cref{eq:rho} anticipates the improvement of \crossmean{} and of the depth-ordered correction, in \Cref{sec:exp-criterion}; and (iii)~test the transfer of \crossmean{} to BERT encoders in \Cref{sec:exp-scope}.

\textbf{Experimental setting.} We merge experts for the standard eight-task CLIP
\citep{radford2021learning} benchmark \citep{ilharco2023editing} and for fourteen- and twenty-task sets that extend it, on five backbones spanning $86$M to $632$M vision parameters. We refer to the twenty tasks of the largest set as our \emph{twenty-task pool}. This yields the fifteen
settings of our shared-protocol grid, namely all the combinations of the five backbones and the three numbers of tasks. In \Cref{tab:main,tab:full}, we merge the publicly released experts of FusionBench \citep{tang2024fusionbench} on the eight-task sets of ViT-B/32, ViT-B/16, and ViT-L/14 with their QuickGELU activation, and we run the shared-protocol grid with GELU, the activation of the experts that we fine-tuned ourselves.  Calibration uses $256$ unlabeled training images per task, and validation uses a further $256$
disjoint images whose labels serve only to select hyperparameters. We evaluate every method on a test
split that is disjoint from the training data of the experts, and no test input enters any merge or any
selection. Every number in our main-text tables is a mean over three calibration seeds, and we define the
margin between two methods as the difference of their test accuracies, in \%. We describe the complete protocol in \Cref{app:protocol}, and we report the cost of each method and the effect of the calibration set size in \Cref{app:cost}.

\textbf{Baselines.} We compare with two groups of methods: (i)~data-free baselines, namely weight
averaging \citep{wortsman2022model}, Task Arithmetic \citep{ilharco2023editing}, DARE
\citep{yu2024dare}, TIES \citep{yadav2023ties}, Model Stock \citep{jang2024modelstock}, and Iso-C
\citep{marczak2025iso}; and (ii)~activation-based baselines, namely RegMean \citep{jin2023regmean} and
Progressive RegMean \citep{buzzega2025chainofmerges}, and, with full coverage, RegMean++
\citep{nguyen2026regmeanpp} and LOT \citep{sun2025lot}. For every method with a scale hyperparameter, we select it on the validation split, as we describe in \Cref{app:protocol,app:lot}. In contrast, RegMean, Progressive RegMean, and \crossmean{} have no scale hyperparameter.

\input{sections/tables}

\subsection{Merging accuracy of \crossmean{}}
\label{sec:exp-correct}

In this Section, we evaluate the merging accuracy of \crossmean{}. Specifically, we (i)~compare it with the baselines under both coverages and apply it to LOT; and (ii)~study the asymmetry of its bilinear statistic.

\textbf{\crossmean{} improves on RegMean on all five backbones.} As we report in
\Cref{tab:main}, with MLP coverage, our \crossmean{} improves on RegMean on all five backbones. Moreover, on our shared-protocol grid,
\crossmean{} achieves a higher accuracy than all the other methods in our comparison in eleven of the fifteen settings, as we
report in \Cref{tab:appfull} of \Cref{app:full}.

\textbf{\crossmean{} improves further with full coverage and when applied to LOT.} Moving from MLP coverage to the
full coverage of \Cref{sec:setup} increases the accuracy of RegMean, Progressive RegMean, and
\crossmean{} by $1.9$ to $3.8\%$ on the eight-task sets of \Cref{tab:full}. In this setting, our
\crossmean{} improves on Progressive RegMean by a further $+0.94$ to $+1.90\%$, in every seed on all
three backbones that we tested. In addition, \crossmean{} applies to another closed-form method. Specifically, we compute the closed-form solution of LOT \citep{sun2025lot} from the bilinear statistic of \crossmean{}, which we denote by LOT + \crossmean{}. Similarly, we denote by LOT + Progressive RegMean the solution of LOT computed from the merged inputs, and we provide the details of both in \Cref{app:lot}. LOT + \crossmean{} improves on LOT on all three backbones, by $+0.50$ to $+3.75\%$.

\textbf{The asymmetry of the bilinear statistic is important.} Unlike a Gram matrix, our bilinear
statistic in \Cref{eq:bilinear} is, in general, not symmetric, because it pairs the merged and the expert
inputs. When we gradually remove its antisymmetric part, the accuracy of \crossmean{} decreases monotonically on both backbones that we tested, whereas shrinking the statistic toward the Gram matrix $\bA$ in \Cref{eq:AC} or dropping calibration tokens changes its margin over Progressive RegMean by less than $1\%$, as we report in
\Cref{tab:ablations} of \Cref{app:ablations}. Hence, the antisymmetric part of the bilinear statistic is important for \crossmean{}.

\subsection{The alignment criterion and the improvement of \crossmean{}}
\label{sec:exp-criterion}

As we show in \Cref{sec:exp-correct}, \crossmean{} improves on RegMean on the eight-task sets of all five backbones. Similarly, Progressive RegMean, the per-layer form of the depth-ordered correction on which \crossmean{} builds, improves on RegMean in all fifteen settings of the shared-protocol grid, as we report in \Cref{tab:margins} of \Cref{app:full}. However, the size of these improvements varies across task sets. In this Section, we show that our
alignment criterion $\alignop$ in \Cref{eq:rho} anticipates this variation. To this end, we (i)~compare, on each of the five backbones, a controlled pair of two disjoint eight-task sets; (ii)~pool the twenty-five task sets of the shared-protocol grid and of the controlled pairs; and (iii)~for the depth-ordered correction, vary only the tasks on a fixed backbone and adjust for task difficulty, the mean accuracy of each expert on its own task.

\input{sections/tables_criterion}

\textbf{On each of the five backbones, \crossmean{} improves more on the task set with the lower
$\alignop$.} Our controlled pairs in \Cref{tab:criterion} use two disjoint eight-task subsets of the twenty-task pool on each backbone, and hence only the tasks differ. On each of the five backbones, the task set with the lower $\alignop$ has the larger margin
of \crossmean{} over RegMean, between $+7.01$ and $+12.50\%$, compared with at most $+1.70\%$ for the
task set with the higher $\alignop$. 
Moreover, on the task sets with the lower $\alignop$, \crossmean{} improves on Progressive RegMean by
$+3.31$ to $+8.26\%$. These results indicate that the bilinear statistic is beneficial in particular
where the merged-prefix mismatch is large.

\textbf{The relation holds across backbones and numbers of tasks.} Next, across the fifteen settings of
the shared-protocol grid and the ten task sets of the controlled pairs, task sets with a lower
$\alignop$ show a larger margin of \crossmean{} over RegMean, with a rank correlation of $-0.72$, as we
show in \Cref{fig:thesis}c. Moreover, over the fifteen settings of the shared-protocol grid, a lower
$\alignop$ is associated with a larger margin of \crossmean{} over Progressive RegMean, with a rank
correlation of $-0.69$. These results suggest that the further the merged inputs depart from the expert
inputs, the more \crossmean{} improves the merge by pairing the two inputs in its bilinear statistic.

\textbf{For the depth-ordered correction, the relation also holds after adjusting for task difficulty.}
Finally, over twelve random eight-task sets drawn from the twenty-task pool on ViT-B/32, with the backbone
and the number of tasks fixed, the rank correlation of $\alignop$ with the margin of Progressive RegMean
over RegMean is $-0.73$. Moreover, it remains $-0.71$ after adjusting for task difficulty, with a $95\%$ interval of $[-0.98,-0.16]$, as we report in
\Cref{tab:uncertainty}. We plot the margin as a function of $\alignop$ and of task difficulty in \Cref{fig:confound} of \Cref{app:confound}.

\subsection{Merging BERT encoders}
\label{sec:exp-scope}

In this Section, we test whether the improvement of \crossmean{} carries over to BERT text encoders. To this end, we merge fine-tuned BERT-base \citep{devlin2019bert} encoders on two GLUE \citep{wang2019glue} task sets of four and six tasks, which we denote by GLUE-4 and GLUE-6, with the same linear system and selection protocol as in our vision experiments. We consider four settings, namely GLUE-4 with $N=256$, $512$, and $2048$ calibration examples per task and GLUE-6 with $N=256$, as we describe in \Cref{app:crossfamily}.

\input{sections/apptab_bert}

\textbf{\crossmean{} improves on RegMean in all four settings.} As we report in \Cref{tab:bertglue}, our
\crossmean{} improves on RegMean in all four settings, by $+5.0$ to $+7.2\%$, while Progressive RegMean scores below RegMean. Moreover, at $N=256$, the task set with the lower $\alignop$, GLUE-6, shows the larger margin of \crossmean{} over RegMean, as in our controlled pairs on the vision backbones. These results suggest that the bilinear statistic is beneficial beyond image encoders.

%% file: sections/tables.tex
\begin{table}[t]
\centering\footnotesize
\setlength{\tabcolsep}{1.9pt}
\caption{Our results on the eight-task sets with the MLP coverage of \Cref{sec:setup}, in test accuracy (\%),
averaged over three calibration seeds. ViT-B/32, B/16, and L/14 use QuickGELU, and the other two
backbones use GELU. We mark the top three in each column as
\rankfirst{First}, \ranksecond{Second}, and \rankthird{Third}.}
\label{tab:main}
\begin{tabular}{lccccc}
\toprule
Method & B/32 & B/16 & L/14 & L/14-336 & H/14 \\
\midrule
\multicolumn{6}{l}{\textsc{Data-free baselines}}\\
\hspace{3pt}Weight averaging & $64.3$ & $70.2$ & $77.9$ & $76.8$ & $82.6$ \\
\hspace{3pt}Task Arithmetic & $67.4$ & $75.2$ & $79.8$ & $83.3$ & $87.6$ \\
\hspace{3pt}DARE & $67.1$ & $75.1$ & $80.1$ & $83.3$ & $87.5$ \\
\hspace{3pt}TIES & $70.6$ & $76.4$ & $82.6$ & $83.7$ & $87.7$ \\
\hspace{3pt}Model Stock & $51.2$ & $57.6$ & $66.4$ & $62.5$ & $71.1$ \\
\hspace{3pt}Iso-C & $62.6$ & $78.0$ & $84.0$ & $77.0$ & $84.4$ \\
\midrule
\multicolumn{6}{l}{\textsc{Activation-based baselines}}\\
\hspace{3pt}RegMean & \rankthird{76.7} & \rankthird{80.5} & \rankthird{86.8} & \rankthird{85.1} & \rankthird{88.9} \\
\hspace{3pt}Progressive RegMean & \ranksecond{79.2} & \ranksecond{82.4} & \ranksecond{87.9} & \rankfirst{86.4} & \ranksecond{89.7} \\
\midrule
\multicolumn{6}{l}{\textsc{Ours}}\\
\hspace{3pt}\crossmean{} & \rankfirst{81.4} & \rankfirst{83.8} & \rankfirst{88.1} & \ranksecond{86.3} & \rankfirst{90.2} \\
\bottomrule
\end{tabular}
\end{table}

\begin{table}[t]
\centering\footnotesize
\setlength{\tabcolsep}{3pt}
\caption{Our results on the eight-task sets with QuickGELU and full coverage, in which all linear
layers are merged. Our \crossmean{} and LOT + \crossmean{} both improve on all the activation-based baselines in this table on all three backbones, including RegMean++ and LOT. Entries are test
accuracies in percent, averaged over three calibration seeds. We mark the top three in each column as
\rankfirst{First}, \ranksecond{Second}, and \rankthird{Third}.}
\label{tab:full}
\begin{tabular}{lccc}
\toprule
Method & B/32 & B/16 & L/14 \\
\midrule
\multicolumn{4}{l}{\textsc{Activation-based baselines}}\\
\hspace{3pt}RegMean & $80.4$ & $84.0$ & $89.1$ \\
\hspace{3pt}RegMean++ & $82.4$ & $85.1$ & \rankthird{89.8} \\
\hspace{3pt}Progressive RegMean & $82.5$ & $85.2$ & \rankthird{89.8} \\
\hspace{3pt}LOT & $80.3$ & $85.0$ & $89.6$ \\
\hspace{3pt}LOT + Progressive RegMean & \rankthird{82.6} & \rankthird{85.3} & $89.6$ \\
\midrule
\multicolumn{4}{l}{\textsc{Ours}}\\
\hspace{3pt}\crossmean{} & \rankfirst{84.4} & \ranksecond{86.7} & \rankfirst{90.8} \\
\hspace{3pt}LOT + \crossmean{} & \ranksecond{84.1} & \rankfirst{87.1} & \ranksecond{90.1} \\
\bottomrule
\end{tabular}
\end{table}

%% file: sections/tables_criterion.tex
\begin{table}[t]
\centering\scriptsize
\setlength{\tabcolsep}{1.3pt}
\caption{Our controlled pairs of task sets. On each of the five backbones, the task set with the lower
$\alignop$ has the larger margins of Progressive RegMean and of our \crossmean{} over RegMean. The two
task sets of each backbone are disjoint eight-task subsets of the twenty-task pool. Margins over RegMean
are in percent, averaged over three calibration seeds, and bold marks the larger margin of each pair.}
\label{tab:criterion}
\begin{tabular}{lcccccc}
\toprule
& \multicolumn{3}{c}{Low-$\alignop$ task set} & \multicolumn{3}{c}{High-$\alignop$ task set} \\
\cmidrule(lr){2-4}\cmidrule(lr){5-7}
Backbone & $\alignop$ & \shortstack{Prog.\\RegMean} & \crossmean{} & $\alignop$ & \shortstack{Prog.\\RegMean} & \crossmean{} \\
\midrule
B/32 & $0.863$ & $\mathbf{+5.81}$ & $\mathbf{+12.50}$ & $0.947$ & $+0.47$ & $+1.70$ \\
B/16 & $0.901$ & $\mathbf{+3.71}$ & $\mathbf{+11.97}$ & $0.963$ & $+0.41$ & $+0.62$ \\
L/14 & $0.934$ & $\mathbf{+3.70}$ & $\mathbf{+7.01}$ & $0.968$ & $+0.40$ & $-0.20$ \\
L/14-336 & $0.950$ & $\mathbf{+5.79}$ & $\mathbf{+10.50}$ & $0.974$ & $+0.40$ & $-0.19$ \\
H/14 & $0.934$ & $\mathbf{+5.02}$ & $\mathbf{+9.49}$ & $0.970$ & $+0.33$ & $+0.55$ \\
\bottomrule
\end{tabular}
\end{table}

%% file: sections/apptab_bert.tex
\begin{table}[t]
\centering\footnotesize
\setlength{\tabcolsep}{2.0pt}
\caption{Our merges of BERT-base encoders on GLUE show that our \crossmean{} improves on RegMean in all four settings. Entries are average GLUE validation scores over three calibration draws, with the margin over RegMean in small type, $N$ is the number of calibration examples per task, and $\alignop$ is our alignment criterion of each setting. We mark the top three in each row as \rankfirst{First},
\ranksecond{Second}, and \rankthird{Third}.}
\label{tab:bertglue}
\begin{tabular}{lrcccc}
\toprule
Task set & $N$ & RegMean & \shortstack{Progressive\\RegMean} & \crossmean{} & $\alignop$ \\
\midrule
GLUE-4 & $256$ & \ranksecond{66.9} & \rankthird{64.0}\,{\scriptsize$-2.98$} & \rankfirst{72.0}\,{\scriptsize$+5.11$} & $0.942$ \\
GLUE-4 & $512$ & \ranksecond{67.0} & \rankthird{64.0}\,{\scriptsize$-2.94$} & \rankfirst{72.0}\,{\scriptsize$+5.04$} & $0.942$ \\
GLUE-4 & $2048$ & \ranksecond{67.0} & \rankthird{64.1}\,{\scriptsize$-2.90$} & \rankfirst{73.3}\,{\scriptsize$+6.29$} & $0.943$ \\
GLUE-6 & $256$ & \ranksecond{66.2} & \rankthird{64.4}\,{\scriptsize$-1.77$} & \rankfirst{73.4}\,{\scriptsize$+7.24$} & $0.936$ \\
\bottomrule
\end{tabular}
\end{table}

%% file: sections/conclusion.tex
\section{Conclusion}
\label{sec:conclusion}

We studied the activation statistics of closed-form merging methods, which are measured on a single
expert although each merged layer receives its inputs from the merged prefix. By formulating these
methods as a linear system, we showed that they use the same Gram matrix on both sides, and by relaxing
this equality, we obtained \crossmean{}, which pairs the merged and the expert inputs. From the same
statistics, we derived a label-free alignment criterion. In our experiments, \crossmean{} improves on RegMean on the eight-task sets of five vision backbones and on BERT encoders, and
with all linear layers merged, it further improves on the depth-ordered correction on each backbone that we tested. Moreover, task sets with a lower value of our
criterion show a larger improvement. Overall, our results suggest that merging benefits from pairing merged and expert activations.

%% file: sections/appendix.tex
\FloatBarrier
\section{Protocol, experts, baselines, and further related work}
\label{app:protocol}

In this Section, we describe (i)~the protocol and the evaluation splits that all methods share; (ii)~the experts and the backbones; (iii)~the details of the merges, of Progressive RegMean, and of the baselines, together with the computing environment; and (iv)~further related work.

\textbf{Settings shared by every method.} We run every method of \Cref{tab:appfull} under the same protocol in all fifteen settings of the shared-protocol grid,
each of which pairs a backbone with a task set. Specifically, for a method with a scale hyperparameter, we select the scale
from the nine-value grid $\lambda\in\{0.1,0.2,0.3,0.4,0.5,0.7,1.0,1.4,2.0\}$ on a held-out
validation split. In contrast, RegMean, Progressive RegMean, and \crossmean{} have no scale hyperparameter, and we use the closed-form solution of \Cref{eq:template} for each of them. We report every method on a disjoint test split, over three calibration seeds,
with a calibration set $\mathcal{D}_t$ of $256$ unlabeled images per task. In addition, RegMean, Progressive RegMean, and \crossmean{} use the stabilization that we describe below.

\textbf{Evaluation splits.} We evaluate every expert on a test split disjoint from its own
training data. The public FusionBench experts \citep{tang2024fusionbench} were fine-tuned on the
training splits of that release. Hence, for SUN397, Stanford Cars, RESISC45, EuroSAT, GTSRB, and DTD, we
use the training split of the release for calibration and validation, and we use its test split for
evaluation. For MNIST and SVHN, we use their official splits, and for the experts that we fine-tuned ourselves, we keep the training and test splits of their fine-tuning. 

\textbf{Experts.} The eight-task sets on ViT-B/32, ViT-B/16, and ViT-L/14 use the public
fine-tuned checkpoints of the FusionBench release \citep{tang2024fusionbench} for the standard benchmark. The fourteen- and
twenty-task sets contain those eight experts, and they add experts that we fine-tuned for five epochs with a learning rate of $10^{-5}$ on at most $20{,}000$ training images per task, following the fine-tuning protocol of the release, which freezes the zero-shot head and trains only the image encoder. Hence, the eight-, fourteen-, and twenty-task sets are nested, and all methods merge the same experts. For the two other backbones, ViT-H/14 and ViT-L/14-336, we fine-tuned experts for all twenty tasks with the same protocol. These two backbones separate two factors, the model size and the number of tokens per image, that vary together among the other three backbones: (i)~ViT-H/14 roughly doubles the parameter
count of ViT-L/14; and (ii)~ViT-L/14-336 keeps the L/14 architecture and increases the number of tokens
per image from $257$ to $577$. The five backbones are ViT-B/32, ViT-B/16, ViT-L/14, ViT-L/14-336, and ViT-H/14, all of them CLIP vision transformers \citep{radford2021learning,dosovitskiy2021image} that we load with OpenCLIP \citep{ilharco2021openclip}. ViT-H/14
uses the weights trained on LAION-2B \citep{schuhmann2022laion,cherti2023reproducible}. In contrast, the other four
backbones use the OpenAI weights. The eight standard tasks are SUN397 \citep{xiao2010sun}, Stanford Cars
\citep{krause20133d}, RESISC45 \citep{cheng2017remote}, EuroSAT \citep{helber2019eurosat}, SVHN
\citep{netzer2011reading}, GTSRB \citep{stallkamp2011german}, MNIST \citep{lecun1998gradient}, and DTD
\citep{cimpoi2014describing}. The fourteen-task sets add CIFAR-10 and CIFAR-100
\citep{krizhevsky2009learning}, STL-10 \citep{coates2011analysis}, Food-101 \citep{bossard2014food},
Oxford-IIIT Pets \citep{parkhi2012cats}, and FGVC-Aircraft \citep{maji2013fine}. The twenty-task
sets further add Fashion-MNIST \citep{xiao2017fashion}, KMNIST \citep{clanuwat2018deep}, EMNIST
\citep{cohen2017emnist}, Flowers-102 \citep{nilsback2008automated}, Rendered SST-2
\citep{socher2013recursive,radford2021learning}, and Country211 \citep{radford2021learning}. 

\textbf{Activation in each setting.} The public experts use the QuickGELU activation, and the experts that we fine-tuned use GELU \citep{hendrycks2016gelu}.
Accordingly, we merge the eight-task sets of ViT-B/32, ViT-B/16, and ViT-L/14 with QuickGELU in
\Cref{tab:main,tab:full,tab:pertask} and in \Cref{app:lot}, and we run the shared-protocol grid with GELU. As we
report in \Cref{tab:activation}, with everything else fixed, the activation changes our criterion $\alignop$ by at most $0.004$ and the margin of Progressive RegMean over RegMean by at most $0.29\%$. In addition, with QuickGELU, \crossmean{} improves on Progressive RegMean on all three backbones.

\input{sections/apptab_activation}

\textbf{Layer coverage.} With both coverages of \Cref{sec:setup}, we average the normalization parameters, the biases, and the embeddings over the experts. With MLP coverage, we solve the linear system for the two linear layers of the MLP in every block, and the attention projections keep
the average of the experts.  In contrast, with full coverage, we also solve the linear systems of the attention input projection and of the attention output projection. The query, key, and value projections share their input. Therefore, we solve for their stacked weights with a shared Gram matrix, which is equivalent to solving for each projection separately.  

\textbf{Stabilization of the linear system.} We scale the ridge \citep{hoerl1970ridge} that stabilizes the linear system by the mean diagonal of the matrix being inverted, rather than adding a fixed absolute value. Because $\bG$ is a Gram matrix, its magnitude increases with both the number of tokens and the scale of the activations. Therefore, an absolute $\epsilon$ would have a different effect at each layer and on each backbone. Specifically, with $\epsilon=10^{-4}$ in every method that inverts a Gram matrix, we use $\big(\bG+\epsilon\,\bar{d}(\bG)\bI\big)^{-1}$, where
$\bar d(\bG)=\tfrac{1}{n}\mathrm{tr}\,\bG$ is the mean diagonal of the $n\times n$ matrix $\bG$.
In addition, we keep the standard stabilization of RegMean, which scales the off-diagonal entries of every Gram matrix by
$0.9$, $\bG\mapsto0.9\,\bG+0.1\,\mathrm{diag}(\bG)$. We call the result the reduced Gram matrix. In RegMean and Progressive RegMean, the same reduced
Gram matrix appears on both sides of \Cref{eq:template}. Therefore, the solution is the minimizer of the
reconstruction objective in which each squared error $\mathrm{tr}(\bD\bG_t\bD^\top)$, with
$\bD=\bW-\bW_t$, is replaced by
$0.9\,\mathrm{tr}(\bD\bG_t\bD^\top)+0.1\sum_i(\bG_t)_{ii}\lVert\bD_{:,i}\rVert^2$. This replacement adds a penalty, weighted by the diagonal of $\bG_t$, that pulls the merged weight toward the weights of each expert. The reduced Gram matrix is a convex combination of two positive
semi-definite matrices, and therefore it remains positive semi-definite. Similarly, \crossmean{} applies the
same reduction to both of its statistics, the Gram matrix $\tilde\bX_t^\top\tilde\bX_t$ of the merged inputs and the bilinear statistic $\bar\bC_t$ of \Cref{eq:bilinear}.

\textbf{Forward passes.} RegMean passes each calibration
set once through its own expert.
By comparison, Progressive RegMean passes every calibration set through the merged prefix once per mergeable layer. Therefore, with $L$ mergeable layers, it needs $LT$ forward passes instead of $T$.
\crossmean{} adds the passes for the points of the set $\mathcal{S}$ of \Cref{sec:crossmean} with $s>0$, which brings its total to at most $3LT$. However, all three methods invert a
matrix of the same size at every layer. We report the measured cost in \Cref{app:cost}.

\textbf{Progressive RegMean.} Progressive RegMean solves the linear system of each mergeable layer in depth order and re-measures the merged inputs $\tilde\bX_t$ of each layer on the merged prefix, so that $\tilde\bX_t$
replaces $\bX_t$ throughout the RegMean objective. This re-measurement adds the forward passes counted in the
previous paragraph and no new hyperparameter over RegMean. Progressive RegMean is the per-layer form of the depth-ordered correction. Like Chain of Merges \citep{buzzega2025chainofmerges}, it re-measures the inputs after every merged linear
layer. In \Cref{app:full}, we relate this form to the block-level re-measurement of RegMean++
\citep{nguyen2026regmeanpp}.

\textbf{Baselines.} Weight averaging \citep{wortsman2022model} and Task Arithmetic
\citep{ilharco2023editing} sum the task vectors of the experts with a global scalar. TIES \citep{yadav2023ties} trims the task vectors, elects a sign for each entry as the sign of the sum of the trimmed task vectors, and averages the entries that agree with this sign. DARE \citep{yu2024dare} drops entries at
random and rescales the remaining entries. Model Stock \citep{jang2024modelstock} interpolates between the average of the
experts and the pretrained weights with a ratio set by the angle between task vectors, and Iso-C
\citep{marczak2025iso} equalizes the spectrum of the summed task vectors. In the linear system in \Cref{eq:template}, all of these methods except Iso-C correspond to $\bA_t=\bB_t=\bI$, up to a mask, an entrywise reweighting, and a scalar. In contrast, Iso-C does not fit this linear system. We run these methods and RegMean \citep{jin2023regmean} as described in their papers. 

\textbf{Computing environment.} We run all merges and evaluations on a single NVIDIA B200. We solve the per-layer linear systems in double precision.   We fix the seeds for sampling the calibration sets and for their order.

\subsection{Further related work}
\label{app:morerelated}

In this Section, we review further work related to our study. Specifically, we discuss (i)~methods that reduce interference between task vectors; (ii)~methods that merge networks trained independently; (iii)~methods that rely on activations, search, or post-hoc correction; and (iv)~measures that compare representations or predict transfer.

\textbf{Interference between task vectors.} Interpolating a fine-tuned model with its zero-shot
initialization improves its robustness \citep{wortsman2022robust}, and a large body of work refines
Task Arithmetic \citep{ilharco2023editing} in weight space. Fine-tuning in the tangent space of the pretrained model
\citep{ortizjimenez2023tangent} improves the disentanglement of task vectors, and $\tau$-Jp
\citep{kato2025mastering} uses the product of a task vector with the Jacobian of the pretrained
model as an indicator of interference. Model Breadcrumbs \citep{davari2023model} and DELLA
\citep{deep2024della} sparsify each task vector before summing, and TALL masks and Consensus Merging
\citep{wang2024localizing} localize the parameters that each task needs. PCB-Merging
\citep{du2024pcbmerging} balances parameter importance within and across tasks, and EMR-Merging
\citep{huang2024emr} keeps lightweight per-task masks and rescalers around a unified task vector.
Concrete subspace learning \citep{tang2023concrete} learns a shared mask, and TATR \citep{sun2025tatr}
restricts the merge to a trust region of parameters within which the losses of the other tasks remain nearly unchanged. TSV-Merge \citep{gargiulo2025tsv} decorrelates the singular vectors of the task matrices,
and LiNeS \citep{wang2025lines} scales the update of each layer with its depth. A study of merging at
scale \citep{yadav2024matters} examines how four merging methods, including Task Arithmetic and TIES, behave as the model size and the number of experts increase. Many of these merging methods act on the weights alone. In addition, tangent-space fine-tuning and $\tau$-Jp change how the experts are fine-tuned, and Concrete subspace learning and the few-shot variant of TATR use input data only to learn a mask or a trust region. In contrast, the methods that we study in this paper solve each layer from its input activations.

\textbf{Merging networks trained independently.} When the experts do not share a pretrained
model, their units should be matched before the experts are merged. For example, OT fusion \citep{singh2020otfusion} aligns neurons with
optimal transport, and Git Re-Basin \citep{ainsworth2023git} finds permutations under which two networks
are approximately linearly mode connected \citep{frankle2020linear}. ZipIt \citep{stoica2024zipit} merges
correlated features within and across models, and REPAIR \citep{jordan2023repair} rescales the hidden
units of the interpolated network to match their statistics to those of the original networks.
REPAIR is related to our work, because it also corrects a mismatch between the activation
statistics of the merged network and those of its endpoints. Distinctively, we study experts that share the same pretrained model. Hence, no matching of units is needed, and the mismatch that we study is the merged-prefix mismatch rather than a mismatch caused by permuted units.

\textbf{Activations, search, and post-hoc correction.} AIM \citep{nobari2025aim} also measures activations on a calibration set, but it uses them to decide which weights of the base model to protect
rather than to solve for the merged weights. Evolutionary model merging \citep{akiba2024evolutionary}
searches over merge recipes, and FuseLLM \citep{wan2024fusellm} fuses language models of different
architectures by training a target model on their output distributions. Representation Surgery
\citep{yang2024surgery} trains a lightweight module on top of the merged model to remove the bias between the representations of the merged model and those of each expert. Evolutionary merging, FuseLLM, and
Representation Surgery require search or training, or add parameters. In contrast, the methods that we study
fit each layer in closed form.

\textbf{Comparing representations and predicting transfer.} Our criterion in \Cref{eq:rho} is a
normalized Frobenius inner product between two statistics of the layer inputs. Similarly, kernel-target alignment
\citep{cristianini2001kernel}, the Hilbert--Schmidt independence criterion
\citep{gretton2005measuring}, and centered kernel alignment
\citep{kornblith2019similarity,debiasedcka2024} compare two representations through inner products of
their Gram matrices. Methods based on canonical correlation analysis \citep{raghu2017svcca,morcos2018insights} and
representational similarity analysis \citep{kriegeskorte2008representational} compare them through
other statistics. These measures compare two representations of the same inputs. In contrast, our criterion $\alignop$ compares the Gram matrix of the merged inputs with the cross statistic of the merged and the expert inputs. In addition, a second line of work predicts how well a pretrained model will transfer to a labeled target task
before fine-tuning it. H-score \citep{bao2019hscore}, LogME \citep{you2021logme}, and TransRate
\citep{huang2022transrate} score how well the features of the pretrained model explain the target
labels. LEEP \citep{nguyen2020leep} scores how well the predictions of that model on the target
data explain the same labels, and NCE \citep{tran2019transferability} scores how well the labels
of a source task on the same inputs explain them. These estimators use the labels of the target task
to rank models or source tasks. Unlike these estimators, our criterion $\alignop$ uses no labels, and we compute it before the closed-form merge.
Moreover, its value is lower for the task sets in which \crossmean{} and the depth-ordered correction improve more on RegMean, as we show in \Cref{sec:exp-criterion}.

\FloatBarrier
\section{Additional merging results}
\label{app:full}

In this Section, we report additional merging results, namely (i)~the accuracy of every method in each setting of the shared-protocol grid; (ii)~the margin of Progressive RegMean over RegMean in each setting; (iii)~the per-task accuracies; (iv)~\crossmean{} and RegMean++ under full coverage; and (v)~\crossmean{} applied to LOT.

We report the accuracy of every method in all fifteen settings of the shared-protocol grid in \Cref{tab:appfull}, in which \crossmean{} achieves a higher accuracy than all the other methods in our comparison in eleven of the fifteen settings. In addition, we report in \Cref{tab:margins} the margin of Progressive RegMean over RegMean in each setting.

We report the per-task accuracies in \Cref{tab:pertask,tab:pertask2}. In the per-task results, on each of the five backbones, \crossmean{} improves on all the baselines on RESISC45, EuroSAT, SVHN, GTSRB, and DTD.  

\input{sections/apptab_full}

\input{sections/tables_margins}

\input{sections/pertask_tables}

\textbf{Full coverage and RegMean++.} RegMean++ \citep{nguyen2026regmeanpp} re-measures the merged inputs only at the boundaries between transformer blocks, and inside a block it keeps the expert inputs. Hence, RegMean++ is a block-level form of the depth-ordered correction, and Progressive RegMean is its per-layer form. As we report in \Cref{tab:full}, with full coverage, the two forms reach accuracies within $0.1\%$ of each other on ViT-B/32, ViT-B/16, and ViT-L/14, and our \crossmean{} improves on both forms on all three backbones. Moreover, the improvement of both forms over RegMean decreases from ViT-B/32 to ViT-L/14, while our criterion $\alignop$ of the task set increases from $0.892$ to $0.928$, as we report in \Cref{tab:activation}.

\subsection{Applying \crossmean{} to LOT Merging}
\label{app:lot}

In this Section, we apply \crossmean{} to LOT Merging \citep{sun2025lot}, which we denote by LOT. To this end, we (i)~describe the protocol; (ii)~define LOT + Progressive RegMean and LOT + \crossmean{}; (iii)~relate LOT to RegMean; and (iv)~report the results. 

\textbf{Protocol.} We run LOT with the same experts, heads, calibration sets of $256$ images per task, and
evaluation as every other method, with QuickGELU, as we describe in \Cref{app:protocol}. We select the
scale $\lambda$ on the validation split over the same nine-value grid as the other methods with a scale hyperparameter. 

\textbf{LOT + Progressive RegMean and LOT + \crossmean{}.} These two methods keep the closed-form
solution of LOT, including its solution for the normalization scales, and change only the statistics from
which LOT computes this solution. Specifically, both methods solve for the parameters in the order that
LOT uses. LOT + Progressive RegMean measures the inputs of each parameter on the merged prefix, and LOT +
\crossmean{} uses the objective in \Cref{eq:crossobj}, expressed in the parametrization of LOT, with the
points $s\in\{0,\tfrac12,1\}$ on the path from the merged prefix to each expert, as discussed in
\Cref{sec:crossmean}. For both methods, we select $\lambda$ on the validation split over
$\{1.0,1.4,2.0\}$.

\input{sections/lot_table}

\textbf{Relation of LOT to RegMean.} LOT \citep{sun2025lot} fits each linear layer so that, with
only that layer merged, it reproduces the features of each expert on the expert inputs, which yields the merged task vector in closed form,
$\Delta^\top=(\sum_t\bG_t)^{\dagger}\sum_t\bG_t\Delta_t^\top$. On the input directions that the calibration
data excite, this closed form coincides with RegMean without the off-diagonal reduction, expressed in task-vector coordinates. Hence, LOT still keeps the same Gram matrix on both sides of \Cref{eq:template}. However, LOT differs from RegMean with MLP coverage in four respects: (i)~it solves the attention input projections; (ii)~it fits normalization scales
in closed form; (iii)~it omits the off-diagonal reduction; and (iv)~it rescales the merged task vector by a selected
factor. With full coverage, which also includes the attention projections, \crossmean{} and Progressive RegMean both exceed LOT on all three backbones, as we report in \Cref{tab:full}.

\textbf{Results.} LOT + \crossmean{} improves on LOT in every seed on all three backbones that we tested, by $+3.75$,
$+2.11$, and $+0.50\%$ on B/32, B/16, and L/14, as we report in \Cref{tab:applot}. Moreover, this improvement
decreases as the $\alignop$ of the task set increases. In addition, on each of the three backbones, LOT + \crossmean{} improves more on LOT than LOT + Progressive RegMean. 

\FloatBarrier
\section{Additional results on the alignment criterion}
\label{app:criterion}

In this Section, we report additional results on the alignment criterion. Specifically, we (i)~list $\alignop$ and the margin of every measured task set, together with the tasks of the controlled pairs; (ii)~relate our criterion to the alignment of Iso-C and to the raw prefix drift, and show its values layer by layer; (iii)~separate our criterion from task difficulty in \Cref{app:confound}; and (iv)~report the uncertainty of every rank correlation in \Cref{app:uncertainty}.

\textbf{All twenty-five task sets.} Across all twenty-five task sets, those with a lower $\alignop$ tend to show a larger margin of \crossmean{} and of Progressive RegMean over RegMean.
We list $\alignop$ and the margin of each task set in \Cref{tab:appalign}.

\input{sections/apptab_align}

\textbf{The controlled pairs.} In the controlled pairs of \Cref{tab:criterion}, the low-$\alignop$ task set consists of MNIST, SVHN, KMNIST, EMNIST, Fashion-MNIST, CIFAR-10, CIFAR-100, and
STL-10, and the high-$\alignop$ task set consists of SUN397, Stanford Cars, RESISC45, EuroSAT, DTD, Food-101,
Flowers-102, and Country211.  

\textbf{Relation to the alignment of Iso-C.} The authors of Iso-C \citep{marczak2025iso} also report that an alignment correlates with merging quality. However, their alignment concerns a different object. Specifically, the alignment of Iso-C measures how
much of each task \emph{matrix} lies in the dominant singular subspace of the merged matrix, and it
correlates with how much merging improves on the pretrained model. In contrast, our criterion compares two statistics of the \emph{activations}, the Gram matrix of the merged inputs and the cross statistic. Moreover, it anticipates the improvements of \crossmean{} and Progressive RegMean over RegMean.

\textbf{On the three eight-task sets, the raw prefix drift does not order the margins.} As discussed in \Cref{sec:criterion}, an alternative to $\alignop$ is the raw prefix drift, $\lVert\tilde\bX_t-\bX_t\rVert_F/\lVert\bX_t\rVert_F$, which we measure from the same weight-averaged model as $\alignop$.
On the eight-task sets of ViT-B/32, B/16, and L/14, the drift is $0.576$, $0.554$, and $0.587$. We measure the drift with GELU, and with GELU, these task sets reach margins of Progressive RegMean over RegMean of $+2.65$, $+1.97$, and $+1.37\%$, and the drift does not change monotonically with these margins. In contrast, over the same task sets, $\alignop$ is $0.894$, $0.919$,
and $0.931$, and it increases as the margins decrease. These results agree with the argument of \Cref{sec:criterion}, which rests on how the two measures weight the input directions. Specifically, with the eigendecomposition $\bA=\sum_i\mu_i\mathbf{u}_i\mathbf{u}_i^\top$, the trace $\mathrm{tr}\,\bC=\sum_i\mathbf{u}_i^\top\bC\mathbf{u}_i$ that enters the drift weights every direction equally, including directions in which the merged inputs carry low energy. In contrast, $\langle\bA,\bC\rangle=\sum_i\mu_i\,\mathbf{u}_i^\top\bC\mathbf{u}_i$ weights each direction by the energy $\mu_i$ that the merged inputs place in it. 

\begin{figure}[ht]
\centering
\includegraphics[width=0.5\textwidth]{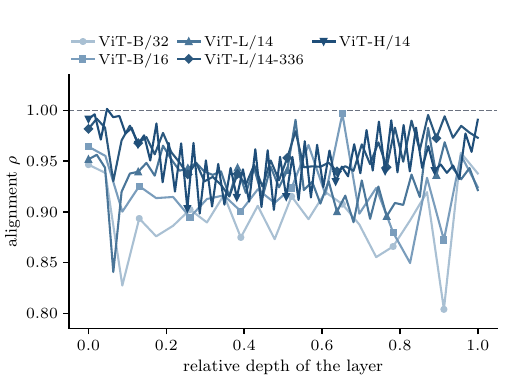}
\caption{We measure the alignment criterion $\alignop$ layer by layer on the eight-task set of each backbone, from the weight-averaged model, and the average of each curve is the criterion of \Cref{sec:criterion}. Our measurements show
that $\alignop$ varies with depth differently on each backbone. The dashed line marks $\alignop=1$, the value of $\alignop$ when the merged inputs equal the expert inputs.}
\label{fig:rhoprofile}
\end{figure}

\textbf{The criterion layer by layer.} As we show in \Cref{fig:rhoprofile}, the value of $\alignop$ at each layer varies with depth differently on each backbone, and the criterion of \Cref{sec:criterion} is the average of these values.

\subsection{Accounting for task difficulty}
\label{app:confound}

\begin{figure}[ht]
\centering
\begin{minipage}[b]{0.46\textwidth}\centering\includegraphics[width=\linewidth]{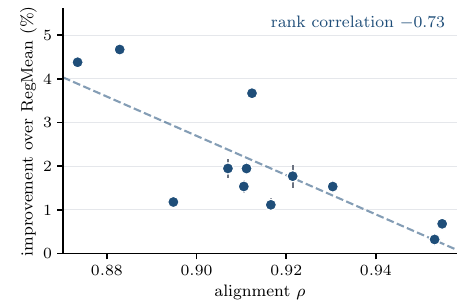}\\[2pt]{\small (a) Margin vs.\ alignment criterion $\alignop$}\end{minipage}\hfill
\begin{minipage}[b]{0.46\textwidth}\centering\includegraphics[width=\linewidth]{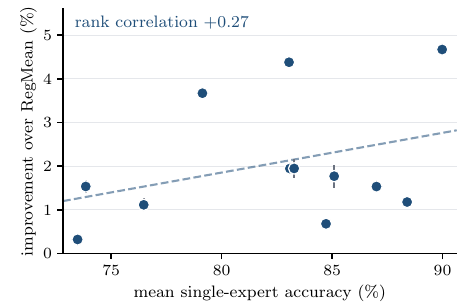}\\[2pt]{\small (b) Margin vs.\ task difficulty}\end{minipage}
\caption{Our merges of twelve random eight-task sets on ViT-B/32, with the backbone and the number of tasks fixed. \textbf{(a)}~Task sets with a lower value of our alignment criterion $\alignop$ show a larger margin of Progressive RegMean over RegMean. \textbf{(b)}~The same margin as a function of task difficulty, namely the mean accuracy of each expert on its own task. Error bars are the standard deviation of the margin over three calibration seeds.}
\label{fig:confound}
\end{figure}

In this Section, we describe the task-difficulty study of \Cref{sec:exp-criterion} and its repetition on ViT-B/16.

\textbf{Design.} Task difficulty is a potential confound. Specifically, task sets whose experts are individually
easier may merge better. Consequently, $\alignop$ could reflect difficulty rather than the merged-prefix mismatch that it is meant to measure.  To separate the two factors, we draw twelve distinct eight-task sets at random, with a fixed seed, from the twenty-task pool on ViT-B/32.  For each task set, we measure (i)~the margin of Progressive RegMean over RegMean, paired within each of three calibration seeds; (ii)~$\alignop$, measured on the weight-averaged model as in \Cref{sec:criterion}; and (iii)~difficulty as the mean accuracy of each expert on
its \emph{own} task. Difficulty measured this way involves no merge, whereas the accuracy of a merged model
would inherit part of the margin.

\textbf{Result.} Over the twelve task sets, whose margins span $+0.32$ to $+4.67\%$, a lower $\alignop$ is associated with
a larger margin, with a rank correlation of $-0.73$. By comparison, the rank correlation of difficulty with the margin is $+0.27$, with
$p=0.41$, and the rank correlation of difficulty with $\alignop$ is $-0.22$. Moreover, after adjusting for difficulty, the partial correlation of $\alignop$ with the margin remains $-0.71$, with an interval
that excludes zero, whereas the partial correlation of difficulty after adjusting for $\alignop$ is $+0.16$.
We compute each partial correlation as the rank correlation of the residuals that remain after a least-squares regression of both ranked variables on the ranks of the adjusted variable.  We plot the margin in \Cref{fig:confound} and report every interval in \Cref{tab:uncertainty}.

\textbf{Repetition on ViT-B/16.} Over the same twelve task sets on ViT-B/16, $\alignop$ and task difficulty keep their signs and their relative strength. Specifically, the rank correlation of $\alignop$ with the margin is $-0.64$
unconditionally and $-0.61$ after adjusting for difficulty, whereas the corresponding correlations of difficulty are $+0.38$ and $+0.33$.
 The unconditional rank correlation of $\alignop$ is significant, with $p=0.033$. 

\textbf{Conclusion of the difficulty study.} With the number of tasks held fixed, our study on ViT-B/32 indicates that our criterion reflects more than task difficulty, and the repetition on ViT-B/16 shows the same signs.

\input{sections/uncertainty_table}

\FloatBarrier
\section{\crossmean{}: design ablations and cost}
\label{app:crossmean}

In this Section, we report (i)~design ablations of \crossmean{}; and (ii)~the cost of \crossmean{} and the effect of the calibration set size.

\subsection{Design ablations}
\label{app:ablations}

\input{sections/apptab_ablations}

In this Section, we change one design choice of \crossmean{} at a time, and we report in \Cref{tab:ablations} the margin of the changed method over Progressive RegMean on the eight-task sets of ViT-B/32 and ViT-L/14 with GELU.

\textbf{The asymmetry of the bilinear statistic is important.} Unlike a Gram matrix, the bilinear statistic is, in general, not symmetric. Removing its antisymmetric part gradually, by using in the linear system the matrix $\bar\bC_\beta=(1-\beta)\bar\bC+\tfrac{\beta}{2}(\bar\bC+\bar\bC^\top)$, lowers the margin monotonically in $\beta$ on both backbones. At $\beta=1$, the margin over Progressive RegMean is $-70.61\%$ on ViT-B/32 and $-78.82\%$ on ViT-L/14.   In contrast, two other modifications of the statistic leave the margin nearly unchanged. First, convex shrinkage toward the Gram matrix $\bA$ that the linear system inverts, $\bar\bC\leftarrow(1-\gamma)\bar\bC+\gamma\bA$, reproduces Progressive RegMean at $\gamma=1$. Second, Bernoulli dropout of calibration tokens at rate $p$ uses the same mask for $\bA$ and $\bar\bC$, and hence the two statistics remain a matched pair. Neither modification changes the sign of the margin on either backbone, and neither modification changes it by more than $0.97\%$. By comparison, symmetrization lowers it by $26$ to $78\%$.

\subsection{Cost and calibration set size}
\label{app:cost}

\input{sections/apptab_cost}

In this Section, we report (i)~the time and memory of one merge; and (ii)~the effect of the calibration set size.

\textbf{Time and memory.} We report the cost of one merge in \Cref{tab:cost}. On the eight-task set of ViT-B/32, \crossmean{} takes $31.2\times$ the time of RegMean, and Progressive RegMean takes $24.4\times$. On ViT-B/32, both ratios remain about the same at twenty tasks, and on ViT-H/14, they are $68.8\times$ for \crossmean{} and $25.9\times$ for Progressive RegMean. Hence, relative to RegMean, the time of \crossmean{} depends on the backbone rather than on the number of tasks. All times include
the forward passes of feature extraction.
In addition, on both backbones in \Cref{tab:cost}, \crossmean{} and Progressive RegMean need $28$ to $34\%$ less peak memory than RegMean, because they do not hold the Gram matrices of all experts at once. 

\begin{figure}[htbp]
\centering
\includegraphics[width=0.5\textwidth]{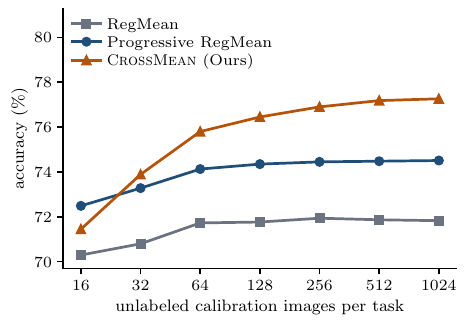}
\caption{We measure the accuracy on the eight-task set of ViT-B/32 with GELU as a function of the number $N$ of calibration images per task. \textcolor{figgray}{RegMean (gray)} and \textcolor{figblue}{Progressive RegMean (blue)} reach a plateau at $N{=}64$, while our \textcolor{figorange}{\crossmean{} (orange)} reaches its plateau at about four times as many images.}
\label{fig:budget}
\end{figure}

\textbf{Calibration set size.} As we show in \Cref{fig:budget}, RegMean and Progressive RegMean reach a plateau at $N{=}64$ unlabeled calibration images per task, while \crossmean{} reaches its plateau at about four times as many images.  From $N{=}32$ onward, \crossmean{} improves on Progressive RegMean. Hence, the $256$ calibration images per task that we use in our vision experiments lie on the plateau of \crossmean{}.

\FloatBarrier
\section{Merging BERT encoders on GLUE}
\label{app:scope}
\label{app:crossfamily}

In this Section, we describe our experiments with BERT encoders in detail. To this end, we (i)~describe the merges and the evaluation; (ii)~describe the calibration draws; and (iii)~report our criterion.

\textbf{Merges and evaluation.} We merge fine-tuned BERT-base encoders on GLUE with the same linear system and selection protocol, over the two task sets GLUE-4 and GLUE-6 of \Cref{sec:exp-scope}. GLUE-4 consists of CoLA, MRPC, RTE, and SST-2, and GLUE-6 adds QNLI and QQP \citep{wang2019glue}. We fine-tune one expert per task from the pretrained BERT-base-uncased encoder \citep{devlin2019bert}, with a learning rate of $2\times10^{-5}$. All methods merge the encoder, including its pooler, and each task keeps the classification head of its expert, which we apply to the pooled output of the merged encoder. We select hyperparameters on a labeled subset of the training split of
each task, disjoint from the calibration set, and we evaluate the selected merge once on the GLUE
validation split. The metric is the Matthews correlation for CoLA and the accuracy for the other tasks, averaged over tasks. We compute each margin over RegMean within a calibration draw, and hence the comparison is paired.

\textbf{Calibration draws.} As in our vision experiments, we draw each calibration set with a fixed seed, and we report the mean over three draws in \Cref{tab:bertglue}. The standard deviation over the three draws is at most $0.35$ for \crossmean{} and $1.21$ for Progressive RegMean.

\textbf{The criterion.} On these encoders, $\alignop$ is $0.9417$ on GLUE-4 and $0.9358$ on GLUE-6, and it changes by at most $0.0011$ across calibration set sizes. Moreover, the task set with the
lower $\alignop$ has the larger improvement of \crossmean{} over RegMean, $+7.24\%$ compared with $+5.11\%$ at $N{=}256$.  

\FloatBarrier

%% file: sections/apptab_activation.tex
\begin{table}[ht]
\centering\footnotesize
\setlength{\tabcolsep}{3.2pt}
\caption{Our comparison of the two activations shows that switching from GELU (G) to QuickGELU (Q) changes our criterion $\alignop$ by at most $0.004$ and the margin of Progressive RegMean over RegMean by at most $0.29\%$, while it increases the accuracy of RegMean and Progressive RegMean by $3.5$ to $4.9\%$. Each row is an eight-task set of public experts, with everything else fixed,
and margins are in percent, averaged over three calibration seeds. The margin of Progressive RegMean is over RegMean, and the margin of \crossmean{} is over Progressive RegMean.}
\label{tab:activation}
\begin{tabular}{lcccccc}
\toprule
& \multicolumn{2}{c}{\shortstack{Progressive\\RegMean margin}} & \multicolumn{2}{c}{\shortstack{\crossmean{}\\margin}} & \multicolumn{2}{c}{$\alignop$} \\
\cmidrule(lr){2-3}\cmidrule(lr){4-5}\cmidrule(lr){6-7}
Backbone & G & Q & G & Q & G & Q \\
\midrule
B/32 & $+2.65$ & $+2.59$ & $+2.84$ & $+2.15$ & $0.894$ & $0.892$ \\
B/16 & $+1.97$ & $+1.92$ & $+1.12$ & $+1.38$ & $0.919$ & $0.915$ \\
L/14 & $+1.37$ & $+1.08$ & $-0.86$ & $+0.19$ & $0.931$ & $0.928$ \\
\bottomrule
\end{tabular}
\end{table}

%% file: sections/apptab_full.tex
\begin{table}[ht]
\centering\footnotesize
\setlength{\tabcolsep}{1.5pt}
\caption{In our experiments, \crossmean{} reaches a higher accuracy than every other method in eleven of the fifteen settings. Entries are test accuracies in percent with GELU and MLP coverage, averaged over three seeds, and the headers indicate the number of tasks $T$. We report the eight-task sets of ViT-B/32, B/16, and L/14 with QuickGELU in \Cref{tab:main}. We mark the top three in each column as \rankfirst{First}, \ranksecond{Second}, and \rankthird{Third}.}
\label{tab:appfull}
\begin{tabular}{lccccccccccccccc}
\toprule
& \multicolumn{3}{c}{B/32} & \multicolumn{3}{c}{B/16} & \multicolumn{3}{c}{L/14}
& \multicolumn{3}{c}{L/14-336} & \multicolumn{3}{c}{H/14} \\
\cmidrule(lr){2-4}\cmidrule(lr){5-7}\cmidrule(lr){8-10}\cmidrule(lr){11-13}\cmidrule(lr){14-16}
Method & 8 & 14 & 20 & 8 & 14 & 20 & 8 & 14 & 20 & 8 & 14 & 20 & 8 & 14 & 20 \\
\midrule
\multicolumn{16}{l}{\textsc{Data-free baselines}}\\
\hspace{3pt}Weight averaging & $58.8$ & $62.7$ & $53.7$ & $65.0$ & $67.8$ & $58.0$ & $73.3$ & $75.6$ & $64.5$ & $76.8$ & $76.6$ & $65.1$ & $82.6$ & $81.8$ & $69.6$ \\
\hspace{3pt}Task Arithmetic & $61.6$ & $63.9$ & $55.3$ & $69.3$ & $70.5$ & $60.0$ & $75.0$ & $77.1$ & $66.4$ & $83.3$ & $80.7$ & $69.0$ & $87.6$ & $84.6$ & $72.4$ \\
\hspace{3pt}DARE & $61.2$ & $63.8$ & $55.2$ & $69.3$ & $70.4$ & $59.9$ & $74.8$ & $77.0$ & $66.7$ & $83.3$ & $80.6$ & $68.9$ & $87.5$ & $84.5$ & $72.4$ \\
\hspace{3pt}TIES & $64.7$ & $65.8$ & $56.3$ & $70.4$ & $71.4$ & $61.0$ & $77.5$ & $78.9$ & $67.5$ & $83.7$ & $81.4$ & $69.4$ & $87.7$ & $84.8$ & $73.0$ \\
\hspace{3pt}Model Stock & $46.8$ & $57.2$ & $50.1$ & $53.9$ & $62.5$ & $54.4$ & $61.0$ & $69.4$ & $59.9$ & $62.5$ & $70.2$ & $60.3$ & $71.1$ & $76.8$ & $66.2$ \\
\hspace{3pt}Iso-C & $57.4$ & $61.8$ & $53.7$ & $73.4$ & $73.8$ & $63.4$ & $78.4$ & $80.4$ & $70.8$ & $77.0$ & $79.3$ & $68.4$ & $84.4$ & $83.7$ & $72.2$ \\
\midrule
\multicolumn{16}{l}{\textsc{Activation-based baselines}}\\
\hspace{3pt}RegMean & \rankthird{71.8} & \rankthird{71.2} & \rankthird{60.5} & \rankthird{76.1} & \rankthird{75.0} & \rankthird{64.2} & \rankthird{83.0} & \ranksecond{82.2} & \rankthird{71.4} & \rankthird{85.1} & \rankthird{83.0} & \rankthird{71.7} & \rankthird{88.9} & \rankthird{86.9} & \rankthird{74.4} \\
\hspace{3pt}Progressive RegMean & \ranksecond{74.5} & \ranksecond{73.1} & \ranksecond{62.6} & \ranksecond{78.1} & \ranksecond{76.9} & \ranksecond{66.1} & \rankfirst{84.4} & \rankfirst{83.3} & \ranksecond{72.7} & \rankfirst{86.4} & \rankfirst{84.5} & \ranksecond{73.8} & \ranksecond{89.7} & \ranksecond{87.8} & \ranksecond{75.8} \\
\midrule
\multicolumn{16}{l}{\textsc{Ours}}\\
\hspace{3pt}\crossmean{} & \rankfirst{77.3} & \rankfirst{75.0} & \rankfirst{65.6} & \rankfirst{79.2} & \rankfirst{77.6} & \rankfirst{68.2} & \ranksecond{83.5} & \rankthird{82.1} & \rankfirst{73.2} & \ranksecond{86.3} & \ranksecond{83.8} & \rankfirst{74.8} & \rankfirst{90.2} & \rankfirst{88.0} & \rankfirst{77.4} \\
\bottomrule
\end{tabular}
\end{table}

%% file: sections/tables_margins.tex
\begin{table}[ht]
\centering\small
\setlength{\tabcolsep}{5pt}
\caption{Our measurements of the margin of Progressive RegMean over RegMean in the fifteen settings of the shared-protocol grid, which our criterion $\alignop$ anticipates in \Cref{sec:exp-criterion}. Each entry is the margin, in percent, averaged over three calibration seeds, with GELU. $T$ is the number of tasks.}
\label{tab:margins}
\begin{tabular}{llccc}
\toprule
Backbone & Parameters & $T{=}8$ & $T{=}14$ & $T{=}20$ \\
\midrule
ViT-B/32 & 88M & $+2.65$ & $+1.85$ & $+2.16$ \\
ViT-B/16 & 86M & $+1.97$ & $+1.95$ & $+1.97$ \\
ViT-L/14 & 304M & $+1.37$ & $+1.08$ & $+1.27$ \\
ViT-L/14-336 & 304M & $+1.26$ & $+1.46$ & $+2.09$ \\
ViT-H/14 & 632M & $+0.80$ & $+0.93$ & $+1.40$ \\
\bottomrule
\end{tabular}
\end{table}

%% file: sections/pertask_tables.tex
\begin{table}[ht]
\centering\footnotesize
\setlength{\tabcolsep}{4pt}
\caption{In our experiments, \crossmean{} improves on every baseline in task average on ViT-B/32, ViT-B/16, and
ViT-L/14. In addition, on all three backbones, it improves on every baseline on RESISC45, EuroSAT, SVHN, GTSRB, and DTD. Entries are test accuracies
in percent per task of the eight-task set, with QuickGELU, averaged over three calibration
seeds. \emph{Avg.} is the average that we report in \Cref{tab:main}.
We mark the top three in each column as \rankfirst{First}, \ranksecond{Second}, and \rankthird{Third}.}
\label{tab:pertask}
\begin{tabular}{lccccccccc}
\toprule
Method & SUN397 & Cars & RESISC45 & EuroSAT & SVHN & GTSRB & MNIST & DTD & Avg. \\
\midrule
\multicolumn{10}{l}{\emph{ViT-B/32}}\\
\multicolumn{10}{l}{\hspace{3pt}\textsc{Data-free baselines}}\\
\hspace{6pt}Weight averaging & $64.3$ & \rankthird{61.2} & $67.1$ & $68.5$ & $64.3$ & $54.8$ & $86.5$ & $47.7$ & $64.3$ \\
\hspace{6pt}Task Arithmetic & $63.3$ & $59.7$ & $67.6$ & $70.7$ & $73.9$ & $62.1$ & $92.9$ & $48.6$ & $67.4$ \\
\hspace{6pt}DARE & $63.3$ & $59.6$ & $67.0$ & $70.0$ & $73.4$ & $62.2$ & $92.6$ & $48.7$ & $67.1$ \\
\hspace{6pt}TIES & $63.5$ & $59.1$ & $69.8$ & $74.2$ & $82.6$ & $73.3$ & $95.5$ & $46.8$ & $70.6$ \\
\hspace{6pt}Model Stock & $62.9$ & $58.5$ & $57.5$ & $54.0$ & $39.0$ & $36.4$ & $57.8$ & $43.3$ & $51.2$ \\
\hspace{6pt}Iso-C & $63.9$ & $53.5$ & $69.9$ & $67.8$ & $67.5$ & $52.2$ & $73.0$ & $52.6$ & $62.6$ \\
\multicolumn{10}{l}{\hspace{3pt}\textsc{Activation-based baselines}}\\
\hspace{6pt}RegMean & \ranksecond{67.6} & \ranksecond{66.0} & \rankthird{79.1} & \rankthird{85.1} & \rankthird{85.7} & \rankthird{75.4} & \rankthird{96.8} & \rankthird{57.5} & \rankthird{76.7} \\
\hspace{6pt}Progressive RegMean & \rankfirst{68.3} & \rankfirst{66.4} & \ranksecond{80.3} & \ranksecond{85.7} & \ranksecond{91.9} & \ranksecond{84.1} & \ranksecond{98.0} & \ranksecond{59.3} & \ranksecond{79.2} \\
\multicolumn{10}{l}{\hspace{3pt}\textsc{Ours}}\\
\hspace{6pt}\crossmean{} & \rankthird{66.6} & \ranksecond{66.0} & \rankfirst{83.0} & \rankfirst{89.1} & \rankfirst{94.5} & \rankfirst{90.2} & \rankfirst{99.0} & \rankfirst{62.8} & \rankfirst{81.4} \\
\midrule
\multicolumn{10}{l}{\emph{ViT-B/16}}\\
\multicolumn{10}{l}{\hspace{3pt}\textsc{Data-free baselines}}\\
\hspace{6pt}Weight averaging & $66.8$ & $68.3$ & $71.0$ & $75.4$ & $75.5$ & $62.5$ & $93.4$ & $48.9$ & $70.2$ \\
\hspace{6pt}Task Arithmetic & $64.8$ & $68.2$ & $70.9$ & $77.3$ & $89.1$ & $81.7$ & $98.2$ & $51.1$ & $75.2$ \\
\hspace{6pt}DARE & $64.2$ & $68.2$ & $70.5$ & $78.7$ & $89.0$ & $81.0$ & $98.2$ & $51.5$ & $75.1$ \\
\hspace{6pt}TIES & $66.3$ & $66.9$ & $74.5$ & $75.5$ & \rankthird{92.3} & \rankthird{85.6} & \rankthird{98.6} & $51.5$ & $76.4$ \\
\hspace{6pt}Model Stock & $63.2$ & $64.4$ & $63.9$ & $53.7$ & $57.0$ & $46.7$ & $67.8$ & $43.9$ & $57.6$ \\
\hspace{6pt}Iso-C & \rankfirst{70.1} & \rankthird{70.9} & \ranksecond{84.2} & $78.9$ & $85.2$ & $79.3$ & $93.8$ & \rankthird{61.9} & $78.0$ \\
\multicolumn{10}{l}{\hspace{3pt}\textsc{Activation-based baselines}}\\
\hspace{6pt}RegMean & $69.6$ & \ranksecond{74.5} & $81.5$ & \rankthird{85.9} & $91.5$ & $82.7$ & $98.1$ & $60.5$ & \rankthird{80.5} \\
\hspace{6pt}Progressive RegMean & \ranksecond{69.9} & \rankfirst{74.6} & \rankthird{82.7} & \ranksecond{87.9} & \ranksecond{94.6} & \ranksecond{88.5} & \ranksecond{98.8} & \ranksecond{62.6} & \ranksecond{82.4} \\
\multicolumn{10}{l}{\hspace{3pt}\textsc{Ours}}\\
\hspace{6pt}\crossmean{} & \rankthird{69.7} & \rankfirst{74.6} & \rankfirst{85.9} & \rankfirst{90.9} & \rankfirst{96.0} & \rankfirst{91.3} & \rankfirst{99.1} & \rankfirst{63.2} & \rankfirst{83.8} \\
\midrule
\multicolumn{10}{l}{\emph{ViT-L/14}}\\
\multicolumn{10}{l}{\hspace{3pt}\textsc{Data-free baselines}}\\
\hspace{6pt}Weight averaging & $71.0$ & $79.9$ & $78.6$ & $81.8$ & $81.2$ & $74.4$ & $96.3$ & $59.7$ & $77.9$ \\
\hspace{6pt}Task Arithmetic & $72.0$ & $79.3$ & $79.8$ & $81.4$ & $86.9$ & $81.9$ & \rankthird{97.5} & $60.1$ & $79.8$ \\
\hspace{6pt}DARE & $72.2$ & $80.1$ & $80.9$ & $82.6$ & $86.3$ & $81.6$ & $97.2$ & $60.3$ & $80.1$ \\
\hspace{6pt}TIES & \rankthird{73.9} & $80.6$ & $83.8$ & $84.0$ & $91.2$ & $88.9$ & \rankthird{97.5} & $60.8$ & $82.6$ \\
\hspace{6pt}Model Stock & $67.9$ & $76.5$ & $67.9$ & $64.6$ & $63.5$ & $53.9$ & $83.8$ & $53.5$ & $66.4$ \\
\hspace{6pt}Iso-C & \rankfirst{74.7} & $84.1$ & \ranksecond{88.6} & $88.9$ & $84.6$ & $87.5$ & $95.1$ & $68.2$ & $84.0$ \\
\multicolumn{10}{l}{\hspace{3pt}\textsc{Activation-based baselines}}\\
\hspace{6pt}RegMean & $73.4$ & \ranksecond{86.3} & $88.0$ & \rankthird{92.9} & \rankthird{94.4} & \rankthird{91.0} & \ranksecond{98.1} & \rankthird{70.7} & \rankthird{86.8} \\
\hspace{6pt}Progressive RegMean & \ranksecond{74.0} & \rankfirst{86.8} & \rankthird{88.5} & \ranksecond{94.4} & \ranksecond{96.0} & \ranksecond{93.3} & \rankfirst{98.9} & \ranksecond{71.5} & \ranksecond{87.9} \\
\multicolumn{10}{l}{\hspace{3pt}\textsc{Ours}}\\
\hspace{6pt}\crossmean{} & $73.5$ & \rankthird{85.8} & \rankfirst{89.5} & \rankfirst{95.3} & \rankfirst{96.4} & \rankfirst{95.2} & $96.0$ & \rankfirst{73.2} & \rankfirst{88.1} \\
\bottomrule
\end{tabular}
\end{table}

\begin{table}[ht]
\centering\footnotesize
\setlength{\tabcolsep}{4pt}
\caption{In our experiments on ViT-L/14-336 and ViT-H/14, with GELU, \crossmean{} again improves on every baseline on
RESISC45, EuroSAT, SVHN, GTSRB, and DTD, and on ViT-H/14, also in task average.
Entries are as in \Cref{tab:pertask}. We mark the top three in each column as \rankfirst{First}, \ranksecond{Second}, and \rankthird{Third}.}
\label{tab:pertask2}
\begin{tabular}{lccccccccc}
\toprule
Method & SUN397 & Cars & RESISC45 & EuroSAT & SVHN & GTSRB & MNIST & DTD & Avg. \\
\midrule
\multicolumn{10}{l}{\emph{ViT-L/14-336}}\\
\multicolumn{10}{l}{\hspace{3pt}\textsc{Data-free baselines}}\\
\hspace{6pt}Weight averaging & $72.4$ & $79.4$ & $75.9$ & $80.6$ & $80.0$ & $72.8$ & $92.5$ & $61.1$ & $76.8$ \\
\hspace{6pt}Task Arithmetic & $74.0$ & $81.5$ & $82.5$ & $90.9$ & $87.6$ & $84.7$ & \rankthird{97.6} & $67.3$ & $83.3$ \\
\hspace{6pt}DARE & $73.8$ & $81.8$ & $82.5$ & $90.6$ & $87.8$ & $84.8$ & \rankthird{97.6} & $67.3$ & $83.3$ \\
\hspace{6pt}TIES & \rankfirst{74.8} & \rankthird{81.9} & $83.7$ & $88.0$ & $87.3$ & \rankthird{88.0} & $97.2$ & \rankthird{68.8} & $83.7$ \\
\hspace{6pt}Model Stock & $69.4$ & $73.8$ & $64.6$ & $52.5$ & $58.3$ & $56.3$ & $70.5$ & $54.5$ & $62.5$ \\
\hspace{6pt}Iso-C & $73.3$ & $75.7$ & $80.3$ & $84.8$ & $75.7$ & $77.3$ & $85.3$ & $63.8$ & $77.0$ \\
\multicolumn{10}{l}{\hspace{3pt}\textsc{Activation-based baselines}}\\
\hspace{6pt}RegMean & \rankthird{74.3} & \rankfirst{83.9} & \rankthird{84.7} & \rankthird{94.3} & \rankthird{91.1} & $87.1$ & $97.5$ & $68.0$ & \rankthird{85.1} \\
\hspace{6pt}Progressive RegMean & \ranksecond{74.5} & \ranksecond{83.8} & \ranksecond{85.2} & \ranksecond{95.4} & \ranksecond{93.3} & \ranksecond{90.9} & \rankfirst{98.7} & \ranksecond{69.1} & \rankfirst{86.4} \\
\multicolumn{10}{l}{\hspace{3pt}\textsc{Ours}}\\
\hspace{6pt}\crossmean{} & $72.4$ & $81.2$ & \rankfirst{85.4} & \rankfirst{96.1} & \rankfirst{93.8} & \rankfirst{92.4} & \ranksecond{98.6} & \rankfirst{70.6} & \ranksecond{86.3} \\
\midrule
\multicolumn{10}{l}{\emph{ViT-H/14}}\\
\multicolumn{10}{l}{\hspace{3pt}\textsc{Data-free baselines}}\\
\hspace{6pt}Weight averaging & $76.8$ & $94.1$ & $79.3$ & $84.7$ & $85.5$ & $76.6$ & $95.6$ & $68.4$ & $82.6$ \\
\hspace{6pt}Task Arithmetic & \rankthird{77.9} & $94.3$ & $86.1$ & $90.8$ & $91.7$ & $88.4$ & $98.2$ & $73.3$ & $87.6$ \\
\hspace{6pt}DARE & $77.8$ & $94.1$ & $86.1$ & $90.4$ & $91.7$ & $88.2$ & $98.2$ & $73.4$ & $87.5$ \\
\hspace{6pt}TIES & \ranksecond{78.1} & $94.3$ & $87.4$ & $89.0$ & $90.8$ & $89.5$ & $97.8$ & \ranksecond{74.8} & $87.7$ \\
\hspace{6pt}Model Stock & $75.9$ & $93.4$ & $70.8$ & $69.4$ & $56.2$ & $60.3$ & $78.7$ & $63.9$ & $71.1$ \\
\hspace{6pt}Iso-C & \rankthird{77.9} & $94.1$ & \ranksecond{87.7} & $82.3$ & $83.9$ & $79.6$ & $95.5$ & $73.9$ & $84.4$ \\
\multicolumn{10}{l}{\hspace{3pt}\textsc{Activation-based baselines}}\\
\hspace{6pt}RegMean & \ranksecond{78.1} & \ranksecond{94.8} & $86.6$ & \rankthird{95.6} & \rankthird{93.1} & \rankthird{90.1} & \rankthird{98.6} & $74.3$ & \rankthird{88.9} \\
\hspace{6pt}Progressive RegMean & \rankfirst{78.3} & \rankfirst{94.9} & \rankthird{87.5} & \ranksecond{96.4} & \ranksecond{94.1} & \ranksecond{92.8} & \ranksecond{99.0} & \rankthird{74.7} & \ranksecond{89.7} \\
\multicolumn{10}{l}{\hspace{3pt}\textsc{Ours}}\\
\hspace{6pt}\crossmean{} & \ranksecond{78.1} & \rankthird{94.7} & \rankfirst{88.6} & \rankfirst{96.5} & \rankfirst{94.7} & \rankfirst{94.4} & \rankfirst{99.1} & \rankfirst{75.5} & \rankfirst{90.2} \\
\bottomrule
\end{tabular}
\end{table}

%% file: sections/lot_table.tex
\begin{table}[ht]
\centering\footnotesize
\setlength{\tabcolsep}{3pt}
\caption{In our experiments, LOT + \crossmean{} improves on LOT on all three backbones, and the improvement
decreases as the alignment criterion $\alignop$ of the task set increases. Entries are test accuracies in percent on the eight-task
sets with QuickGELU, averaged over three calibration seeds.  We mark the top three in each column as \rankfirst{First}, \ranksecond{Second}, and \rankthird{Third}. In the last row, we report $\alignop$ of each task set.}
\label{tab:applot}
\begin{tabular}{lccc}
\toprule
 & ViT-B/32 & ViT-B/16 & ViT-L/14 \\
\midrule
\multicolumn{4}{l}{\textsc{Baselines}} \\
\hspace{3pt}LOT & \rankthird{80.32} & \rankthird{85.01} & \ranksecond{89.64} \\
\midrule
\multicolumn{4}{l}{\textsc{Applied to LOT}} \\
\hspace{3pt}LOT + Progressive RegMean & \ranksecond{82.61} & \ranksecond{85.26} & \rankthird{89.60} \\
\hspace{3pt}LOT + \crossmean{} & \rankfirst{84.07} & \rankfirst{87.12} & \rankfirst{90.14} \\
\midrule
\multicolumn{4}{l}{\textsc{Improvement over LOT (\%)}} \\
\hspace{3pt}LOT + Progressive RegMean & $+2.29$ & $+0.25$ & $-0.04$ \\
\hspace{3pt}LOT + \crossmean{} & $+3.75$ & $+2.11$ & $+0.50$ \\
\midrule
\multicolumn{4}{l}{\textsc{Alignment criterion}} \\
\hspace{3pt}$\alignop$ of the task set & $0.892$ & $0.915$ & $0.928$ \\
\bottomrule
\end{tabular}
\end{table}

%% file: sections/apptab_align.tex
\begin{table}[ht]
\centering\small
\setlength{\tabcolsep}{6pt}
\caption{Our measurements show that a lower value of the alignment criterion $\alignop$ is associated with a larger margin of \crossmean{} and of Progressive RegMean over RegMean, in percent. Over these task sets, the rank correlations are $-0.72$ and $-0.73$, respectively, and we report their intervals in \Cref{tab:uncertainty}. The rows are the fifteen settings of the shared-protocol grid and the ten task sets of the controlled pairs, sorted by $\alignop$.}
\label{tab:appalign}
\begin{tabular}{lccc}
\toprule
Task set & $\alignop$ & \shortstack{Progressive\\RegMean} & \crossmean{} \\
\midrule
ViT-B/32, low-$\alignop$ task set & $0.8635$ & $+5.81$ & $+12.50$ \\
ViT-B/32, $T{=}20$ & $0.8801$ & $+2.16$ & $+5.12$ \\
ViT-B/32, $T{=}8$ & $0.8943$ & $+2.65$ & $+5.49$ \\
ViT-B/16, low-$\alignop$ task set & $0.9006$ & $+3.71$ & $+11.97$ \\
ViT-B/32, $T{=}14$ & $0.9101$ & $+1.85$ & $+3.83$ \\
ViT-B/16, $T{=}20$ & $0.9105$ & $+1.97$ & $+4.03$ \\
ViT-B/16, $T{=}8$ & $0.9190$ & $+1.97$ & $+3.08$ \\
ViT-L/14, $T{=}8$ & $0.9307$ & $+1.37$ & $+0.51$ \\
ViT-L/14, low-$\alignop$ task set & $0.9336$ & $+3.70$ & $+7.01$ \\
ViT-B/16, $T{=}14$ & $0.9339$ & $+1.95$ & $+2.67$ \\
ViT-H/14, low-$\alignop$ task set & $0.9343$ & $+5.02$ & $+9.49$ \\
ViT-L/14, $T{=}20$ & $0.9364$ & $+1.27$ & $+1.72$ \\
ViT-L/14, $T{=}14$ & $0.9438$ & $+1.08$ & $-0.05$ \\
ViT-H/14, $T{=}20$ & $0.9449$ & $+1.40$ & $+3.06$ \\
ViT-B/32, high-$\alignop$ task set & $0.9468$ & $+0.47$ & $+1.70$ \\
ViT-L/14-336, low-$\alignop$ task set & $0.9501$ & $+5.79$ & $+10.50$ \\
ViT-H/14, $T{=}8$ & $0.9517$ & $+0.80$ & $+1.31$ \\
ViT-L/14-336, $T{=}20$ & $0.9544$ & $+2.09$ & $+3.09$ \\
ViT-H/14, $T{=}14$ & $0.9556$ & $+0.93$ & $+1.12$ \\
ViT-L/14-336, $T{=}8$ & $0.9575$ & $+1.26$ & $+1.22$ \\
ViT-L/14-336, $T{=}14$ & $0.9613$ & $+1.46$ & $+0.72$ \\
ViT-B/16, high-$\alignop$ task set & $0.9631$ & $+0.41$ & $+0.62$ \\
ViT-L/14, high-$\alignop$ task set & $0.9677$ & $+0.40$ & $-0.20$ \\
ViT-H/14, high-$\alignop$ task set & $0.9701$ & $+0.33$ & $+0.55$ \\
ViT-L/14-336, high-$\alignop$ task set & $0.9737$ & $+0.40$ & $-0.19$ \\
\bottomrule
\end{tabular}
\end{table}

%% file: sections/uncertainty_table.tex
\subsection{Uncertainty of the rank correlations}
\label{app:uncertainty}

In this Section, we report in \Cref{tab:uncertainty} every rank correlation that we quote in this paper, with a percentile-bootstrap $95\%$
interval \citep{efron1979bootstrap} over task sets and a $p$-value under a two-sided permutation null. We compute both from $10^4$ resamples at a
fixed seed. 

\begin{table}[ht]
\centering\footnotesize
\setlength{\tabcolsep}{6pt}
\caption{We report every rank correlation that we quote in this paper, with its percentile-bootstrap $95\%$ interval over
task sets and its two-sided permutation $p$-value, both computed from $10^4$ resamples. Unless stated otherwise, the margin is that of Progressive RegMean over RegMean. }
\label{tab:uncertainty}
\begin{tabular}{lccc}
\toprule
Quantity & $n$ & Est. & $95\%$ CI~[$p$] \\
\midrule
$\alignop$ vs.\ margin of \crossmean{}, all 25 measured task sets & 25 & $-0.72$ & $[-0.91,-0.40]$ ~[$2\times10^{-4}$] \\
$\alignop$ vs.\ margin of \crossmean{}, shared-protocol grid & 15 & $-0.68$ & $[-0.95,-0.17]$ ~[$0.006$] \\
$\alignop$ vs.\ margin of \crossmean{} over Progressive RegMean, shared-protocol grid & 15 & $-0.69$ & $[-0.96,-0.19]$ ~[$0.006$] \\
$\alignop$ vs.\ margin of Progressive RegMean, all 25 measured task sets & 25 & $-0.73$ & $[-0.90,-0.43]$ ~[$10^{-4}$] \\
$\alignop$ vs.\ margin, ViT-B/32 confound study & 12 & $-0.73$ & $[-0.94,-0.20]$ ~[$0.010$] \\
Difficulty vs.\ margin, ViT-B/32 confound study & 12 & $+0.27$ & $[-0.50,+0.79]$ ~[$0.408$] \\
$\alignop$ vs.\ difficulty, ViT-B/32 confound study & 12 & $-0.22$ & $[-0.80,+0.48]$ ~[$0.493$] \\
Partial $\alignop$ vs.\ margin given difficulty, ViT-B/32 & 12 & $-0.71$ & $[-0.98,-0.16]$ ~[$0.017$] \\
Partial difficulty vs.\ margin given $\alignop$, ViT-B/32 & 12 & $+0.16$ & $[-0.55,+0.90]$ ~[$0.635$] \\
$\alignop$ vs.\ margin, ViT-B/16 confound study & 12 & $-0.64$ & $[-0.92,-0.04]$ ~[$0.033$] \\
Difficulty vs.\ margin, ViT-B/16 confound study & 12 & $+0.38$ & $[-0.34,+0.89]$ ~[$0.221$] \\
$\alignop$ vs.\ difficulty, ViT-B/16 confound study & 12 & $-0.22$ & $[-0.78,+0.55]$ ~[$0.501$] \\
Partial $\alignop$ vs.\ margin given difficulty, ViT-B/16 & 12 & $-0.61$ & $[-0.94,+0.22]$ ~[$0.050$] \\
Partial difficulty vs.\ margin given $\alignop$, ViT-B/16 & 12 & $+0.33$ & $[-0.37,+0.88]$ ~[$0.319$] \\
\bottomrule
\end{tabular}
\end{table}

\textbf{Implications of the intervals.} The intervals support three conclusions: (i)~in the
ViT-B/32 confound study, $\alignop$ remains associated with the margin after adjusting for difficulty, with an
interval that excludes zero, while neither correlation of difficulty with the margin is significant;
(ii)~the ViT-B/16 repetition has the same signs, and its unconditional correlation is also
significant; and (iii)~both pooled correlations over all twenty-five task sets, for \crossmean{} and for Progressive RegMean, have intervals that exclude zero and $p$-values of at most $2\times10^{-4}$.

%% file: sections/apptab_ablations.tex
\begin{table}[ht]
\centering\small
\setlength{\tabcolsep}{5pt}
\caption{Our ablations show that, of the changes that we tested, only symmetrizing the bilinear statistic lowers the margin of \crossmean{} over
Progressive RegMean by more than $1\%$. Each entry is that margin, in percent, with
one design choice changed, on the eight-task sets with GELU.  Here, $\beta$ is the weight of the symmetric part of the statistic, $\gamma$ is the weight of the shrinkage toward the Gram matrix of the merged inputs, and $p$ is the dropout rate of the calibration tokens. We define each change in \Cref{app:ablations}.}
\label{tab:ablations}
\begin{tabular}{lcc}
\toprule
Change to \crossmean{} & ViT-B/32 & ViT-L/14 \\
\midrule
None, the unmodified method & $+2.84$ & $-0.86$ \\
\midrule
\multicolumn{3}{l}{\textsc{Statistic}} \\
\hspace{3pt}$\beta=0.25$ & $-25.92$ & $-27.20$ \\
\hspace{3pt}$\beta=0.50$ & $-61.46$ & $-78.55$ \\
\hspace{3pt}$\beta=1.00$ & $-70.61$ & $-78.82$ \\
\hspace{3pt}$\gamma=0.15$ & $+2.40$ & $-0.62$ \\
\hspace{3pt}$\gamma=0.30$ & $+2.22$ & $-1.02$ \\
\hspace{3pt}$\gamma=0.50$ & $+1.87$ & $-0.34$ \\
\hspace{3pt}$p=0.10$ & $+2.50$ & $-0.42$ \\
\hspace{3pt}$p=0.25$ & $+2.46$ & $-0.51$ \\
\hspace{3pt}$p=0.40$ & $+2.40$ & $-0.43$ \\
\bottomrule
\end{tabular}
\end{table}

%% file: sections/apptab_cost.tex
\begin{table}[ht]
\centering\small
\setlength{\tabcolsep}{4pt}
\caption{Our measurements of the cost of one merge show that \crossmean{} and Progressive RegMean need less peak memory than RegMean. On one NVIDIA B200, we measure (i)~the wall-clock time of one merge relative to RegMean; and (ii)~the peak allocated memory.}
\label{tab:cost}
\begin{tabular}{lccccc}
\toprule
& \multicolumn{3}{c}{Time relative to RegMean} & \multicolumn{2}{c}{Peak memory (GiB)} \\
\cmidrule(lr){2-4}\cmidrule(lr){5-6}
Method & B/32, $T{=}8$ & B/32, $T{=}20$ & H/14, $T{=}8$ & B/32, $T{=}8$ & H/14, $T{=}8$ \\
\midrule
Iso-C & $0.3\times$ & $0.1\times$ & $0.6\times$ & $0.34$ & $2.36$ \\
RegMean & $1.0\times$ & $1.0\times$ & $1.0\times$ & $1.34$ & $9.50$ \\
Progressive RegMean & $24.4\times$ & $26.8\times$ & $25.9\times$ & $0.96$ & $6.37$ \\
\crossmean{} & $31.2\times$ & $32.4\times$ & $68.8\times$ & $0.92$ & $6.27$ \\
\bottomrule
\end{tabular}
\end{table}

%% file: references.bib
@inproceedings{dosovitskiy2021image,
  title={An Image is Worth 16x16 Words: Transformers for Image Recognition at Scale},
  author={Dosovitskiy, Alexey and Beyer, Lucas and Kolesnikov, Alexander and Weissenborn, Dirk and Zhai, Xiaohua and Unterthiner, Thomas and Dehghani, Mostafa and Minderer, Matthias and Heigold, Georg and Gelly, Sylvain and Uszkoreit, Jakob and Houlsby, Neil},
  booktitle={International Conference on Learning Representations},
  year={2021}
}

@article{tang2024fusionbench,
  title={{FusionBench}: A Unified Library and Comprehensive Benchmark for Deep Model Fusion},
  author={Tang, Anke and Shen, Li and Luo, Yong and Yang, Enneng and Hu, Han and Zhang, Lefei and Du, Bo and Tao, Dacheng},
  journal={Journal of Machine Learning Research},
  volume={26},
  number={307},
  year={2025}
}

@article{akiba2024evolutionary,
  title={Evolutionary Optimization of Model Merging Recipes},
  author={Akiba, Takuya and Shing, Makoto and Tang, Yujin and Sun, Qi and Ha, David},
  journal={Nature Machine Intelligence},
  volume={7},
  number={2},
  year={2025}
}

@inproceedings{ainsworth2023git,
  title={{Git Re-Basin}: Merging Models modulo Permutation Symmetries},
  author={Ainsworth, Samuel K. and Hayase, Jonathan and Srinivasa, Siddhartha},
  booktitle={International Conference on Learning Representations},
  year={2023}
}

@inproceedings{sun2025tatr,
  title={Task Arithmetic in Trust Region: A Training-Free Model Merging Approach to Navigate Knowledge Conflicts},
  author={Sun, Wenju and Li, Qingyong and Wang, Wen and Geng, Yangliao and Li, Boyang},
  booktitle={Proceedings of the 33rd ACM International Conference on Multimedia},
  year={2025}
}

@inproceedings{jordan2023repair,
  title={{REPAIR}: {RE}normalizing Permuted Activations for Interpolation Repair},
  author={Jordan, Keller and Sedghi, Hanie and Saukh, Olga and Entezari, Rahim and Neyshabur, Behnam},
  booktitle={International Conference on Learning Representations},
  year={2023}
}

@inproceedings{wortsman2022robust,
  title={Robust fine-tuning of zero-shot models},
  author={Wortsman, Mitchell and Ilharco, Gabriel and Kim, Jong Wook and Li, Mike and Kornblith, Simon and Roelofs, Rebecca and Gontijo Lopes, Raphael and Hajishirzi, Hannaneh and Farhadi, Ali and Namkoong, Hongseok and Schmidt, Ludwig},
  booktitle={Proceedings of the IEEE/CVF Conference on Computer Vision and Pattern Recognition},
  year={2022}
}

@inproceedings{davari2023model,
  title={Model Breadcrumbs: Scaling Multi-task Model Merging with Sparse Masks},
  author={Davari, MohammadReza and Belilovsky, Eugene},
  booktitle={European Conference on Computer Vision (ECCV)},
  series={Lecture Notes in Computer Science},
  volume={15133},
  publisher={Springer},
  year={2024}
}

@article{tang2023concrete,
  title={Concrete Subspace Learning based Interference Elimination for Multi-task Model Fusion},
  author={Tang, Anke and Shen, Li and Luo, Yong and Ding, Liang and Hu, Han and Du, Bo and Tao, Dacheng},
  journal={arXiv preprint arXiv:2312.06173v1},
  year={2023}
}

@inproceedings{cristianini2001kernel,
  title={On kernel-target alignment},
  author={Cristianini, Nello and Shawe-Taylor, John and Elisseeff, Andr{\'e} and Kandola, Jaz},
  booktitle={Advances in Neural Information Processing Systems},
  volume={14},
  year={2001}
}

@inproceedings{gretton2005measuring,
  title={Measuring Statistical Dependence with {Hilbert-Schmidt} Norms},
  author={Gretton, Arthur and Bousquet, Olivier and Smola, Alex and Sch{\"o}lkopf, Bernhard},
  booktitle={International Conference on Algorithmic Learning Theory},
  series={Lecture Notes in Computer Science},
  volume={3734},
  publisher={Springer},
  doi={10.1007/11564089_7},
  year={2005}
}

@inproceedings{kornblith2019similarity,
  title={Similarity of Neural Network Representations Revisited},
  author={Kornblith, Simon and Norouzi, Mohammad and Lee, Honglak and Hinton, Geoffrey},
  booktitle={Proceedings of the 36th International Conference on Machine Learning},
  series={Proceedings of Machine Learning Research},
  volume={97},
  publisher={PMLR},
  year={2019}
}

@article{kriegeskorte2008representational,
  title={Representational similarity analysis -- connecting the branches of systems neuroscience},
  author={Kriegeskorte, Nikolaus and Mur, Marieke and Bandettini, Peter A.},
  journal={Frontiers in Systems Neuroscience},
  volume={2},
  year={2008}
}

@inproceedings{ilharco2023editing,
  title={Editing models with task arithmetic},
  author={Ilharco, Gabriel and Ribeiro, Marco Tulio and Wortsman, Mitchell and Gururangan, Suchin and Schmidt, Ludwig and Hajishirzi, Hannaneh and Farhadi, Ali},
  booktitle={International Conference on Learning Representations},
  year={2023}
}

@inproceedings{yadav2023ties,
  title={{TIES-Merging}: Resolving Interference When Merging Models},
  author={Yadav, Prateek and Tam, Derek and Choshen, Leshem and Raffel, Colin and Bansal, Mohit},
  booktitle={Advances in Neural Information Processing Systems},
  volume={36},
  year={2023}
}

@inproceedings{yu2024dare,
  title={Language Models are Super {M}ario: Absorbing Abilities from Homologous Models as a Free Lunch},
  author={Yu, Le and Yu, Bowen and Yu, Haiyang and Huang, Fei and Li, Yongbin},
  booktitle={International Conference on Machine Learning},
  series={Proceedings of Machine Learning Research},
  volume={235},
  year={2024}
}

@inproceedings{matena2022merging,
  title={Merging models with {Fisher}-weighted averaging},
  author={Matena, Michael S. and Raffel, Colin},
  booktitle={Advances in Neural Information Processing Systems},
  volume={35},
  year={2022}
}

@inproceedings{jin2023regmean,
  title={Dataless Knowledge Fusion by Merging Weights of Language Models},
  author={Jin, Xisen and Ren, Xiang and Preotiuc-Pietro, Daniel and Cheng, Pengxiang},
  booktitle={International Conference on Learning Representations},
  year={2023}
}

@inproceedings{yang2024adamerging,
  title={{AdaMerging}: Adaptive model merging for multi-task learning},
  author={Yang, Enneng and Wang, Zhenyi and Shen, Li and Liu, Shiwei and Guo, Guibing and Wang, Xingwei and Tao, Dacheng},
  booktitle={International Conference on Learning Representations},
  year={2024}
}

@inproceedings{acemerging2025,
  title={{ACE-Merging}: Data-Free Model Merging with Adaptive Covariance Estimation},
  author={Xu, Bo and Wu, Haotian and Lin, Hehai and Huang, Weiquan and Zhu, Beier and Shu, Yao and Qin, Chengwei},
  booktitle={Proceedings of the IEEE/CVF Conference on Computer Vision and Pattern Recognition},
  year={2026}
}

@inproceedings{wortsman2022model,
  title={Model soups: averaging weights of multiple fine-tuned models improves accuracy without increasing inference time},
  author={Wortsman, Mitchell and Ilharco, Gabriel and Gadre, Samir Ya and Roelofs, Rebecca and Gontijo-Lopes, Raphael and Morcos, Ari S. and Namkoong, Hongseok and Farhadi, Ali and Carmon, Yair and Kornblith, Simon and Schmidt, Ludwig},
  booktitle={International Conference on Machine Learning},
  series={Proceedings of Machine Learning Research},
  volume={162},
  year={2022}
}

@inproceedings{you2021logme,
  title={{LogME}: Practical Assessment of Pre-trained Models for Transfer Learning},
  author={You, Kaichao and Liu, Yong and Wang, Jianmin and Long, Mingsheng},
  booktitle={Proceedings of the 38th International Conference on Machine Learning},
  series={Proceedings of Machine Learning Research},
  volume={139},
  publisher={PMLR},
  year={2021}
}

@inproceedings{bao2019hscore,
  title={An Information-Theoretic Approach to Transferability in Task Transfer Learning},
  author={Bao, Yajie and Li, Yang and Huang, Shao-Lun and Zhang, Lin and Zheng, Lizhong and Zamir, Amir and Guibas, Leonidas},
  booktitle={IEEE International Conference on Image Processing (ICIP)},
  year={2019}
}

@inproceedings{raghu2017svcca,
  title={{SVCCA}: Singular Vector Canonical Correlation Analysis for Deep Learning Dynamics and Interpretability},
  author={Raghu, Maithra and Gilmer, Justin and Yosinski, Jason and Sohl-Dickstein, Jascha},
  booktitle={Advances in Neural Information Processing Systems},
  volume={30},
  year={2017}
}

@inproceedings{morcos2018insights,
  title={Insights on representational similarity in neural networks with canonical correlation},
  author={Morcos, Ari and Raghu, Maithra and Bengio, Samy},
  booktitle={Advances in Neural Information Processing Systems},
  volume={31},
  year={2018}
}

@inproceedings{radford2021learning,
  title={Learning Transferable Visual Models From Natural Language Supervision},
  author={Radford, Alec and Kim, Jong Wook and Hallacy, Chris and Ramesh, Aditya and Goh, Gabriel and Agarwal, Sandhini and Sastry, Girish and Askell, Amanda and Mishkin, Pamela and Clark, Jack and Krueger, Gretchen and Sutskever, Ilya},
  booktitle={Proceedings of the 38th International Conference on Machine Learning},
  series={Proceedings of Machine Learning Research},
  volume={139},
  publisher={PMLR},
  year={2021}
}

@inproceedings{wang2019glue,
  title={{GLUE}: A Multi-Task Benchmark and Analysis Platform for Natural Language Understanding},
  author={Wang, Alex and Singh, Amanpreet and Michael, Julian and Hill, Felix and Levy, Omer and Bowman, Samuel R.},
  booktitle={International Conference on Learning Representations},
  year={2019}
}

@article{debiasedcka2024,
  title={Correcting Biased Centered Kernel Alignment Measures in Biological and Artificial Neural Networks},
  author={Murphy, Alex and Zylberberg, Joel and Fyshe, Alona},
  journal={arXiv preprint arXiv:2405.01012},
  year={2024}
}

@inproceedings{nobari2025aim,
  title={Activation-informed merging of large language models},
  author={Heyrani Nobari, Amin and Alimohammadi, Kaveh and ArjomandBigdeli, Ali and Srivastava, Akash and Ahmed, Faez and Azizan, Navid},
  booktitle={Advances in Neural Information Processing Systems},
  volume={38},
  year={2025}
}

@inproceedings{yang2024surgery,
  title={Representation Surgery for Multi-Task Model Merging},
  author={Yang, Enneng and Shen, Li and Wang, Zhenyi and Guo, Guibing and Chen, Xiaojun and Wang, Xingwei and Tao, Dacheng},
  booktitle={International Conference on Machine Learning},
  series={Proceedings of Machine Learning Research},
  volume={235},
  year={2024}
}

@inproceedings{stoica2024zipit,
  title={{ZipIt!} Merging Models from Different Tasks without Training},
  author={Stoica, George and Bolya, Daniel and Bjorner, Jakob and Ramesh, Pratik and Hearn, Taylor and Hoffman, Judy},
  booktitle={International Conference on Learning Representations},
  year={2024}
}

@article{deep2024della,
  title={{DELLA}-Merging: Reducing Interference in Model Merging through Magnitude-Based Sampling},
  author={Deep, Pala Tej and Bhardwaj, Rishabh and Poria, Soujanya},
  journal={arXiv preprint arXiv:2406.11617},
  year={2024}
}

@inproceedings{kato2025mastering,
  title={Mastering Task Arithmetic: $\tau${Jp} as a Key Indicator for Weight Disentanglement},
  author={Yoshida, Kotaro and Naraki, Yuji and Horie, Takafumi and Yamaki, Ryosuke and Shimizu, Ryotaro and Saito, Yuki and McAuley, Julian and Naganuma, Hiroki},
  booktitle={International Conference on Learning Representations},
  year={2025}
}

@inproceedings{devlin2019bert,
  title={{BERT}: Pre-training of Deep Bidirectional Transformers for Language Understanding},
  author={Devlin, Jacob and Chang, Ming-Wei and Lee, Kenton and Toutanova, Kristina},
  booktitle={Proceedings of the 2019 Conference of the North {A}merican Chapter of the Association for Computational Linguistics: Human Language Technologies, Volume 1 (Long and Short Papers)},
  publisher={Association for Computational Linguistics},
  doi={10.18653/v1/N19-1423},
  year={2019}
}

@inproceedings{huang2024emr,
  title={{EMR}-Merging: Tuning-Free High-Performance Model Merging},
  author={Huang, Chenyu and Ye, Peng and Chen, Tao and He, Tong and Yue, Xiangyu and Ouyang, Wanli},
  booktitle={Advances in Neural Information Processing Systems},
  volume={37},
  year={2024}
}

@inproceedings{nguyen2020leep,
  title={{LEEP}: A New Measure to Evaluate Transferability of Learned Representations},
  author={Nguyen, Cuong V. and Hassner, Tal and Seeger, Matthias and Archambeau, Cedric},
  booktitle={International Conference on Machine Learning},
  series={Proceedings of Machine Learning Research},
  volume={119},
  year={2020}
}

@inproceedings{tran2019transferability,
  title={Transferability and Hardness of Supervised Classification Tasks},
  author={Tran, Anh T. and Nguyen, Cuong V. and Hassner, Tal},
  booktitle={Proceedings of the IEEE/CVF International Conference on Computer Vision},
  year={2019}
}

@inproceedings{huang2022transrate,
  title={Frustratingly Easy Transferability Estimation},
  author={Huang, Long-Kai and Huang, Junzhou and Rong, Yu and Yang, Qiang and Wei, Ying},
  booktitle={International Conference on Machine Learning},
  series={Proceedings of Machine Learning Research},
  volume={162},
  year={2022}
}

@inproceedings{wang2024localizing,
  title={Localizing Task Information for Improved Model Merging and Compression},
  author={Wang, Ke and Dimitriadis, Nikolaos and Ortiz-Jimenez, Guillermo and Fleuret, Fran{\c{c}}ois and Frossard, Pascal},
  booktitle={International Conference on Machine Learning},
  series={Proceedings of Machine Learning Research},
  volume={235},
  year={2024}
}

@article{buzzega2025chainofmerges,
  title={Rethinking Layer-wise Model Merging through Chain of Merges},
  author={Buzzega, Pietro and Salami, Riccardo and Porrello, Angelo and Calderara, Simone},
  journal={arXiv preprint arXiv:2508.21421},
  year={2025}
}

@inproceedings{lu2024twinmerging,
  title={{Twin-Merging}: Dynamic Integration of Modular Expertise in Model Merging},
  author={Lu, Zhenyi and Fan, Chenghao and Wei, Wei and Qu, Xiaoye and Chen, Dangyang and Cheng, Yu},
  booktitle={Advances in Neural Information Processing Systems},
  volume={37},
  year={2024}
}

@inproceedings{du2024pcbmerging,
  title={Parameter Competition Balancing for Model Merging},
  author={Du, Guodong and Lee, Junlin and Li, Jing and Jiang, Runhua and Guo, Yifei and Yu, Shuyang and Liu, Hanting and Goh, Sim Kuan and Tang, Ho-Kin and He, Daojing and Zhang, Min},
  booktitle={Advances in Neural Information Processing Systems},
  volume={37},
  year={2024}
}

@article{shen2024wemoe,
  title={Efficient and Effective Weight-Ensembling Mixture of Experts for Multi-Task Model Merging},
  author={Shen, Li and Tang, Anke and Yang, Enneng and Guo, Guibing and Luo, Yong and Zhang, Lefei and Cao, Xiaochun and Du, Bo and Tao, Dacheng},
  journal={IEEE Transactions on Pattern Analysis and Machine Intelligence},
  volume={48},
  number={3},
  year={2026}
}

@inproceedings{ortizjimenez2023tangent,
  title={Task Arithmetic in the Tangent Space: Improved Editing of Pre-Trained Models},
  author={Ortiz-Jimenez, Guillermo and Favero, Alessandro and Frossard, Pascal},
  booktitle={Advances in Neural Information Processing Systems},
  volume={36},
  year={2023}
}

@article{yang2024mergingsurvey,
  title={Model Merging in {LLMs}, {MLLMs}, and Beyond: Methods, Theories, Applications, and Opportunities},
  author={Yang, Enneng and Shen, Li and Guo, Guibing and Wang, Xingwei and Cao, Xiaochun and Zhang, Jie and Tao, Dacheng},
  journal={ACM Computing Surveys},
  volume={58},
  number={8},
  year={2026}
}

@inproceedings{jang2024modelstock,
  title={Model Stock: All We Need Is Just a Few Fine-Tuned Models},
  author={Jang, Dong-Hwan and Yun, Sangdoo and Han, Dongyoon},
  booktitle={European Conference on Computer Vision},
  year={2024}
}

@inproceedings{wan2024fusellm,
  title={Knowledge Fusion of Large Language Models},
  author={Wan, Fanqi and Huang, Xinting and Cai, Deng and Quan, Xiaojun and Bi, Wei and Shi, Shuming},
  booktitle={International Conference on Learning Representations},
  year={2024}
}

@article{nguyen2026regmeanpp,
  title={{RegMean++}: Enhancing Effectiveness and Generalization of Regression Mean for Model Merging},
  author={Nguyen, The-Hai and Dang, Huu-Tien and Suzuki, Takeshi and Nguyen, Le-Minh},
  journal={Transactions on Machine Learning Research},
  year={2026}
}

@inproceedings{marczak2025iso,
  title={No Task Left Behind: Isotropic Model Merging with Common and Task-Specific Subspaces},
  author={Marczak, Daniel and Magistri, Simone and Cygert, Sebastian and Twardowski, Bart{\l}omiej and Bagdanov, Andrew D. and Van De Weijer, Joost},
  booktitle={International Conference on Machine Learning},
  series={Proceedings of Machine Learning Research},
  volume={267},
  year={2025}
}

@inproceedings{sun2025lot,
  title={Towards Minimizing Feature Drift in Model Merging: Layer-wise Task Vector Fusion for Adaptive Knowledge Integration},
  author={Sun, Wenju and Li, Qingyong and Wang, Wen and Liu, Yang and Geng, Yangliao and Li, Boyang},
  booktitle={Advances in Neural Information Processing Systems},
  volume={38},
  year={2025}
}

@inproceedings{xu2025prodistill,
  title={Scalable Model Merging with Progressive Layer-wise Distillation},
  author={Xu, Jing and Li, Jiazheng and Zhang, Jingzhao},
  booktitle={International Conference on Machine Learning},
  series={Proceedings of Machine Learning Research},
  volume={267},
  year={2025}
}

@inproceedings{xiao2010sun,
  title={{SUN} database: Large-scale scene recognition from abbey to zoo},
  author={Xiao, Jianxiong and Hays, James and Ehinger, Krista A. and Oliva, Aude and Torralba, Antonio},
  booktitle={IEEE Computer Society Conference on Computer Vision and Pattern Recognition},
  year={2010}
}

@inproceedings{krause20133d,
  title={{3D} Object Representations for Fine-Grained Categorization},
  author={Krause, Jonathan and Stark, Michael and Deng, Jia and Fei-Fei, Li},
  booktitle={IEEE International Conference on Computer Vision Workshops},
  year={2013}
}

@article{cheng2017remote,
  title={Remote Sensing Image Scene Classification: Benchmark and State of the Art},
  author={Cheng, Gong and Han, Junwei and Lu, Xiaoqiang},
  journal={Proceedings of the {IEEE}},
  volume={105},
  number={10},
  year={2017}
}

@article{helber2019eurosat,
  title={{EuroSAT}: A Novel Dataset and Deep Learning Benchmark for Land Use and Land Cover Classification},
  author={Helber, Patrick and Bischke, Benjamin and Dengel, Andreas and Borth, Damian},
  journal={{IEEE} Journal of Selected Topics in Applied Earth Observations and Remote Sensing},
  volume={12},
  number={7},
  year={2019}
}

@inproceedings{netzer2011reading,
  title={Reading Digits in Natural Images with Unsupervised Feature Learning},
  author={Netzer, Yuval and Wang, Tao and Coates, Adam and Bissacco, Alessandro and Wu, Bo and Ng, Andrew Y.},
  booktitle={{NIPS} Workshop on Deep Learning and Unsupervised Feature Learning},
  year={2011}
}

@inproceedings{stallkamp2011german,
  title={The {German} Traffic Sign Recognition Benchmark: A multi-class classification competition},
  author={Stallkamp, Johannes and Schlipsing, Marc and Salmen, Jan and Igel, Christian},
  booktitle={The 2011 International Joint Conference on Neural Networks},
  publisher={IEEE},
  doi={10.1109/IJCNN.2011.6033395},
  year={2011}
}

@article{lecun1998gradient,
  title={Gradient-based learning applied to document recognition},
  author={LeCun, Yann and Bottou, L{\'e}on and Bengio, Yoshua and Haffner, Patrick},
  journal={Proceedings of the {IEEE}},
  volume={86},
  number={11},
  year={1998}
}

@inproceedings{cimpoi2014describing,
  title={Describing Textures in the Wild},
  author={Cimpoi, Mircea and Maji, Subhransu and Kokkinos, Iasonas and Mohamed, Sammy and Vedaldi, Andrea},
  booktitle={IEEE Conference on Computer Vision and Pattern Recognition},
  year={2014}
}

@techreport{krizhevsky2009learning,
  title={Learning Multiple Layers of Features from Tiny Images},
  author={Krizhevsky, Alex},
  institution={University of Toronto},
  type={Technical Report},
  year={2009}
}

@inproceedings{coates2011analysis,
  title={An Analysis of Single-Layer Networks in Unsupervised Feature Learning},
  author={Coates, Adam and Ng, Andrew and Lee, Honglak},
  booktitle={International Conference on Artificial Intelligence and Statistics},
  series={Proceedings of Machine Learning Research},
  volume={15},
  year={2011}
}

@inproceedings{bossard2014food,
  title={Food-101 -- Mining Discriminative Components with Random Forests},
  author={Bossard, Lukas and Guillaumin, Matthieu and Van Gool, Luc},
  booktitle={European Conference on Computer Vision},
  year={2014}
}

@inproceedings{parkhi2012cats,
  title={Cats and dogs},
  author={Parkhi, Omkar M. and Vedaldi, Andrea and Zisserman, Andrew and Jawahar, C. V.},
  booktitle={IEEE Conference on Computer Vision and Pattern Recognition},
  year={2012}
}

@misc{maji2013fine,
  title={Fine-Grained Visual Classification of Aircraft},
  author={Maji, Subhransu and Rahtu, Esa and Kannala, Juho and Blaschko, Matthew and Vedaldi, Andrea},
  howpublished={arXiv preprint arXiv:1306.5151},
  eprint={1306.5151},
  archivePrefix={arXiv},
  year={2013}
}

@misc{xiao2017fashion,
  title={{Fashion-MNIST}: a Novel Image Dataset for Benchmarking Machine Learning Algorithms},
  author={Xiao, Han and Rasul, Kashif and Vollgraf, Roland},
  howpublished={arXiv preprint arXiv:1708.07747},
  eprint={1708.07747},
  archivePrefix={arXiv},
  year={2017}
}

@misc{clanuwat2018deep,
  title={Deep Learning for Classical {Japanese} Literature},
  author={Clanuwat, Tarin and Bober-Irizar, Mikel and Kitamoto, Asanobu and Lamb, Alex and Yamamoto, Kazuaki and Ha, David},
  howpublished={arXiv preprint arXiv:1812.01718},
  eprint={1812.01718},
  archivePrefix={arXiv},
  year={2018}
}

@inproceedings{cohen2017emnist,
  title={{EMNIST}: Extending {MNIST} to handwritten letters},
  author={Cohen, Gregory and Afshar, Saeed and Tapson, Jonathan and van Schaik, Andr{\'e}},
  booktitle={International Joint Conference on Neural Networks},
  year={2017}
}

@inproceedings{nilsback2008automated,
  title={Automated Flower Classification over a Large Number of Classes},
  author={Nilsback, Maria-Elena and Zisserman, Andrew},
  booktitle={Sixth Indian Conference on Computer Vision, Graphics \& Image Processing},
  year={2008}
}

@inproceedings{socher2013recursive,
  title={Recursive Deep Models for Semantic Compositionality Over a Sentiment Treebank},
  author={Socher, Richard and Perelygin, Alex and Wu, Jean and Chuang, Jason and Manning, Christopher D. and Ng, Andrew and Potts, Christopher},
  booktitle={Proceedings of the 2013 Conference on Empirical Methods in Natural Language Processing},
  year={2013}
}

@misc{ilharco2021openclip,
  title={{OpenCLIP}},
  author={Ilharco, Gabriel and Wortsman, Mitchell and Carlini, Nicholas and Taori, Rohan and Dave, Achal and Shankar, Vaishaal and Namkoong, Hongseok and Miller, John and Hajishirzi, Hannaneh and Farhadi, Ali and Schmidt, Ludwig},
  howpublished={Zenodo, version 0.1},
  doi={10.5281/zenodo.5143773},
  year={2021}
}

@inproceedings{cherti2023reproducible,
  title={Reproducible Scaling Laws for Contrastive Language-Image Learning},
  author={Cherti, Mehdi and Beaumont, Romain and Wightman, Ross and Wortsman, Mitchell and Ilharco, Gabriel and Gordon, Cade and Schuhmann, Christoph and Schmidt, Ludwig and Jitsev, Jenia},
  booktitle={Proceedings of the IEEE/CVF Conference on Computer Vision and Pattern Recognition},
  year={2023}
}

@inproceedings{schuhmann2022laion,
  title={{LAION-5B}: An open large-scale dataset for training next generation image-text models},
  author={Schuhmann, Christoph and Beaumont, Romain and Vencu, Richard and Gordon, Cade and Wightman, Ross and Cherti, Mehdi and Coombes, Theo and Katta, Aarush and Mullis, Clayton and Wortsman, Mitchell and Schramowski, Patrick and Kundurthy, Srivatsa and Crowson, Katherine and Schmidt, Ludwig and Kaczmarczyk, Robert and Jitsev, Jenia},
  booktitle={Advances in Neural Information Processing Systems (Datasets and Benchmarks Track)},
  volume={35},
  year={2022}
}

@misc{hendrycks2016gelu,
  title={Gaussian Error Linear Units ({GELUs})},
  author={Hendrycks, Dan and Gimpel, Kevin},
  howpublished={arXiv preprint arXiv:1606.08415},
  eprint={1606.08415},
  archivePrefix={arXiv},
  year={2016}
}

@article{hoerl1970ridge,
  title={Ridge Regression: Biased Estimation for Nonorthogonal Problems},
  author={Hoerl, Arthur E. and Kennard, Robert W.},
  journal={Technometrics},
  volume={12},
  number={1},
  year={1970}
}

@article{efron1979bootstrap,
  title={Bootstrap Methods: Another Look at the Jackknife},
  author={Efron, Bradley},
  journal={The Annals of Statistics},
  volume={7},
  number={1},
  year={1979}
}

@article{tam2024mats,
  title={Merging by Matching Models in Task Parameter Subspaces},
  author={Tam, Derek and Bansal, Mohit and Raffel, Colin},
  journal={Transactions on Machine Learning Research},
  year={2024}
}

@inproceedings{cheng2025wudi,
  title={Whoever Started the Interference Should End It: Guiding Data-Free Model Merging via Task Vectors},
  author={Cheng, Runxi and Xiong, Feng and Wei, Yongxian and Zhu, Wanyun and Yuan, Chun},
  booktitle={International Conference on Machine Learning},
  series={Proceedings of Machine Learning Research},
  volume={267},
  year={2025}
}

@inproceedings{gargiulo2025tsv,
  title={Task Singular Vectors: Reducing Task Interference in Model Merging},
  author={Gargiulo, Antonio Andrea and Crisostomi, Donato and Bucarelli, Maria Sofia and Scardapane, Simone and Silvestri, Fabrizio and Rodol{\`a}, Emanuele},
  booktitle={Proceedings of the IEEE/CVF Conference on Computer Vision and Pattern Recognition},
  year={2025}
}

@inproceedings{singh2020otfusion,
  title={Model Fusion via Optimal Transport},
  author={Singh, Sidak Pal and Jaggi, Martin},
  booktitle={Advances in Neural Information Processing Systems},
  volume={33},
  year={2020}
}

@inproceedings{frankle2020linear,
  title={Linear Mode Connectivity and the Lottery Ticket Hypothesis},
  author={Frankle, Jonathan and Dziugaite, Gintare Karolina and Roy, Daniel M. and Carbin, Michael},
  booktitle={International Conference on Machine Learning},
  series={Proceedings of Machine Learning Research},
  volume={119},
  year={2020}
}

@inproceedings{wang2025lines,
  title={{LiNeS}: Post-training Layer Scaling Prevents Forgetting and Enhances Model Merging},
  author={Wang, Ke and Dimitriadis, Nikos and Favero, Alessandro and Ortiz-Jimenez, Guillermo and Fleuret, Fran{\c{c}}ois and Frossard, Pascal},
  booktitle={International Conference on Learning Representations},
  year={2025}
}

@article{yadav2024matters,
  title={What Matters for Model Merging at Scale?},
  author={Yadav, Prateek and Vu, Tu and Lai, Jonathan and Chronopoulou, Alexandra and Faruqui, Manaal and Bansal, Mohit and Munkhdalai, Tsendsuren},
  journal={Transactions on Machine Learning Research},
  year={2025}
}
